\documentclass{article}

    \usepackage[preprint, nonatbib]{neurips_2024}

\usepackage[utf8]{inputenc} 
\usepackage[T1]{fontenc}    
\usepackage{hyperref}       
\usepackage{url}            
\usepackage{booktabs}       
\usepackage{amsfonts}       
\usepackage{nicefrac}       
\usepackage{microtype}      
\usepackage{xcolor}         

\usepackage[pdftex]{graphicx}
\usepackage{booktabs} 
\usepackage{multirow} 
\usepackage{marvosym} 

\newcommand{\simulator}{\texttt{MineLand}}
\newcommand{\simulatorRaw}{{MineLand}} 
\newcommand{\agentRaw}{{Alex}}

\definecolor{drj}{RGB}{67,151,143}
\definecolor{fjq}{RGB}{221,160,221}

\newcommand{\agent}{\texttt{Alex}}

\title{\simulatorRaw: Simulating Large-Scale Multi-Agent Interactions with Limited Multimodal Senses and Physical Needs}

\author{
    Xianhao Yu~\thanks{Equal contribution. The order of authors was determined randomly.}\,\,$^1$ \hspace{9mm}
    Jiaqi Fu~\footnotemark[1]\,\,$^1$ \hspace{9mm}
    Renjia Deng~\footnotemark[1]\,\,$^1$ \hspace{9mm}
    Wenjuan Han~$^{\textrm{\Letter}1}$
    \AND \textnormal{$^1$Beijing Jiaotong University} \\
    {\tt \href{mailto:wjhan@bjtu.edu.cn}{wjhan@bjtu.edu.cn}}
}

\begin{document}

\maketitle

\begin{abstract}
While Vision-Language Models (VLMs) hold promise for tasks requiring extensive collaboration, traditional multi-agent simulators have facilitated rich explorations of an interactive artificial society that reflects collective behavior. However, these existing simulators face significant limitations. Firstly, they struggle with handling large numbers of agents due to high resource demands. Secondly, they often assume agents possess perfect information and limitless capabilities, hindering the ecological validity of simulated social interactions. To bridge this gap, we propose a multi-agent Minecraft simulator, \simulator, that bridges this gap by introducing three key features: large-scale scalability, limited multimodal senses, and physical needs. Our simulator supports 64 or more agents. Agents have limited visual, auditory, and environmental awareness, forcing them to actively communicate and collaborate to fulfill physical needs like food and resources. Additionally, we further introduce an AI agent framework, \agent, inspired by multitasking theory, enabling agents to handle intricate coordination and scheduling. Our experiments demonstrate that the simulator, the corresponding benchmark, and the AI agent framework contribute to more ecological and nuanced collective behavior.
The source code of MineLand and Alex is openly available at https://github.com/cocacola-lab/MineLand.

\end{abstract}

\begin{figure}[ht]
    \centering
    \includegraphics[width=\linewidth]{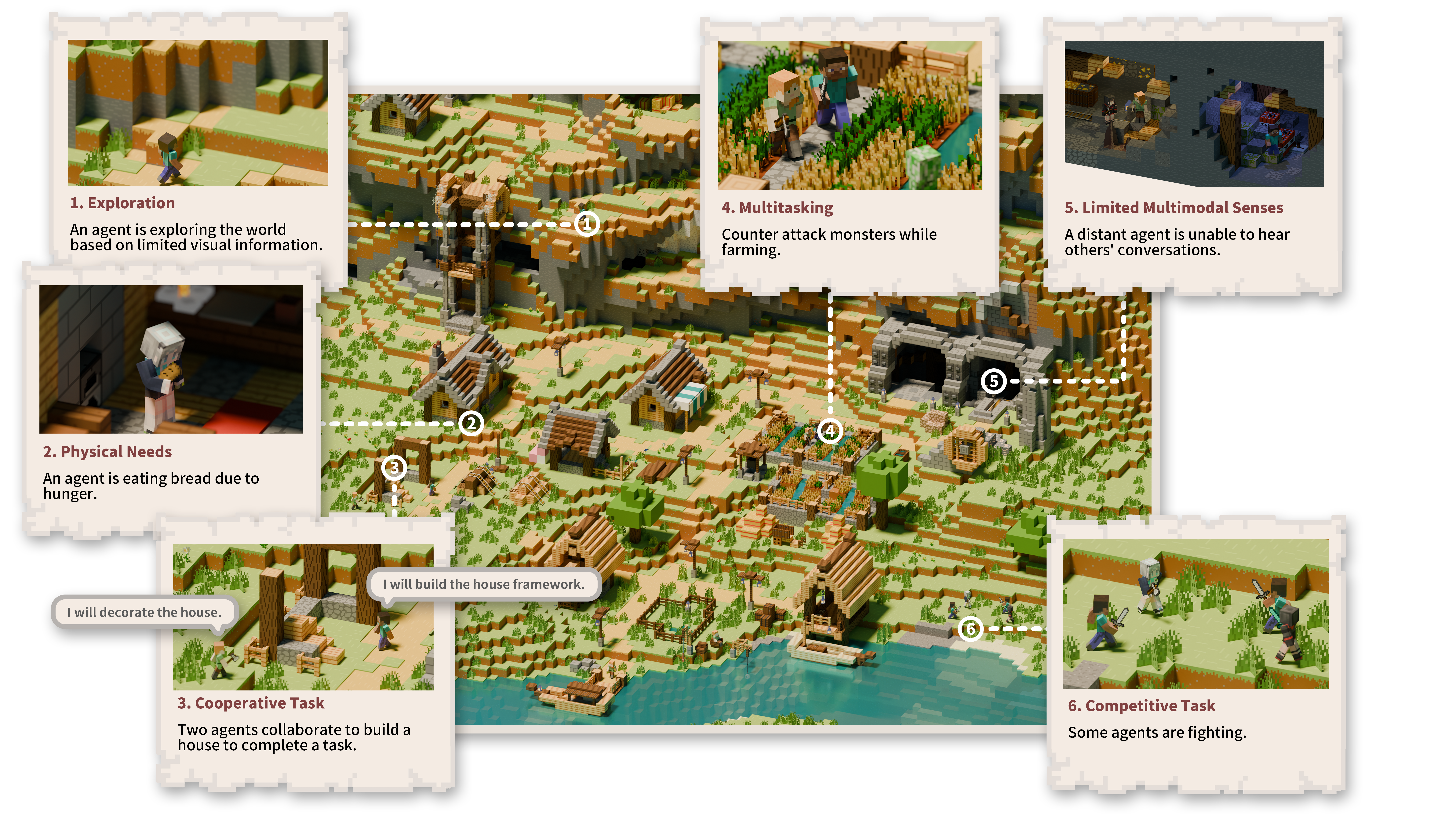}
    \caption{A panoramic view of one scene in \simulator, consisting of multiple AI agents. Subfigure 3\&6 show interactions demonstrating cooperation and competition among several agents. Subfigure 5\&2\&4 showcases the scenarios where the limited senses, physical needs, and multitasking mechanism reflect. In Subfigure 1, an agent is performing a creative task named \textit{Exploration}. Two agents in the left cave of Subfigure 5 cooperate to finish a programmatic \textit{mining} task, while agents in Subfigure 3 are carrying out \textit{building construction}, which is a hybrid task.}
    \label{fig:motivation}
    \vspace{-5mm}
\end{figure}

\newcommand{\hiddenfootnote}[2][\thefootnote]{%
  \footnotetext[#1]{#2}%
}
\renewcommand{\thefootnote}{\textrm{\Letter}}
\hiddenfootnote[1]{Corresponding author.}
\renewcommand{\thefootnote}{\arabic{footnote}}
\setcounter{footnote}{0}

\section{Introduction}\label{sec:intro}

Multi-agent simulators have facilitated rich explorations of an interactive artificial society that reflects collective behavior. From sandbox games such as Smallville~\cite{park2023generative} to virtual environments~\cite{22_bates1994role,4_laird2001human,gong2023mindagent}, researchers and practitioners have been building open-world simulators that can carry multi-agent behaviors and navigate complex human relationships for decades. Especially with the advent of Large Language Models (LLMs) and Vision-Language Models (VLMs), numerous multi-agent simulators based on these technologies have been at the forefront in various fields, from fundamental research to practical applications, such as watch-and-help (WAH) task~\cite{zhang2023building}, Smallville~\cite{park2023generative} and Overcook games~\cite{gong2023mindagent}. However, conventional multi-agent open-world simulators, valuable for exploring collective behavior, suffer from limitations (detailed comparison in Table \ref{table:comparison_of_platforms}). Firstly, they struggle with large-scale scenarios due to the enormous resource consumption required for large-scale agents. Secondly, they are often under the assumption of perfect information and limitless capabilities. These idealized worlds diverge sharply from the messy reality of human interaction. This gap between existing simulators and the real world hinders the ecological validity and richness of social interaction~\cite{heil1983perception}. 

To bridge this gap, we propose \simulator~(Section \ref{sec:simulator}), a multi-agent Minecraft simulator, as shown in Figure~\ref{fig:motivation}, by introducing three key features: large-scale scalability, limited multimodal senses, and physical needs.
First and foremost, the essence of the \simulator's features is the ability to handle the maximum number of agents. Compared to two-agent WAH, single-agent MineDojo~\cite{fan2022minedojo} and twenty-five-agent Smallville~\cite{park2023generative}, our \simulator~enables the utilization of sixty-four or even more agents on the mainstream consumer desktop PC. Secondly, our simulator operates under the human-like assumption~\cite{heil1983perception} that agents possess only limited multimodal senses: partially observable environments, eco-centric perspective, and limited visual and auditory senses. This mirrors real-life social interactions, where visibility and audibility can be affected by factors such as distance, terrain, and environment. These limitations restrict information access, forcing agents to actively communicate to compensate for sensory deficiencies.
Thirdly, we integrate realistic physical needs into agents. Agents require fundamental physical needs, such as food, sustenance, and resource management, which adds a time-based aspect to their daily routine procedures. This necessitates collaboration and competition for resources, mirroring the complex interplay of cooperation and self-interest observed in human societies~\cite{doyal1984theory,alderfer1969empirical}.
By incorporating these three features, our simulator fosters the emergence of dynamic and ecologically valid multi-agent interactions\footnote{Ecological validity refers to interactions between agents within a simulated environment that closely resemble real-world human interactions. For example, actions are situated, adaptive, and environmentally constrained.}.

As an open-world multi-agent simulator, \simulator~is an excellent platform for benchmarking multi-agent capabilities (Section \ref{sec:benchmark}), hence we crowd-sourced datasets to fully evaluate the potential of LLM- or VLM-based multi-agents. For previously existing tasks,  programmatic tasks (e.g., harvest, tech tree, combat tasks) and creative tasks (e.g., survival tasks), we offer a significantly expanded task quantity. Specifically, we crowd-sourced 4499 programmatic tasks and 1536 creative tasks, which is 2 times compared to MineDojo. Interestingly, we introduce a novel ``hybrid task'' category that combines the features of programmatic tasks and creative tasks. Construction and Stage Performance tasks exemplify this new category, bringing the total to 18 hybrid tasks. Additionally, the simulator allows customization of the number of agents and facilitates exploration through two distinct modes: cooperative and competitive.


In addition, we design an AI agent framework - \agent~- inspired by Multitasking theory from the Cognition field~\cite{salvucci2008threaded} (Section \ref{sec:agent}). \agent~allows for simultaneously executing intricate coordination and scheduling with multiple tasks. With this AI agent framework, we have obtained the following intriguing findings: 
(1) Our simulator shows the feature of large-scale scalability (supporting 64 agents, x8 times than before, ($\S$\ref{subsec:expt_performance}), limited multimodal senses ($\S$\ref{subsec:expt_limitated_senses}) and physical needs ($\S$\ref{subsec:expt_physical_needs}). By incorporating these three features, our simulator fosters social dynamics ($\S$\ref{subsec:expt_social_dynamics}\&$\S$\ref{sec:exp_simulating_sociological_phenomena}).
(2) Our benchmark and datasets are challenging ($\S$\ref{subsec:exp_construction_task}\&$\S$\ref{subsec:expt_stage_performance}).
(3) Our agents work together more effectively, with a reduced workload per agent ($\S$\ref{subsec:expt_multi_agent}), and the multitasking mechanism for agents is beneficial ($\S$\ref{subsec:expt_multitasking}).

In summary, our contributions are three-fold: the simulator, benchmark, and AI agent. With these contributions, we push the boundaries of multi-agent simulation by bridging the gap between virtual agents and large-scale real-world humans. This not only advances understanding of AI multi-agents but also holds potential for applications in human dynamics, social psychology, robotics, and game design.
We anticipate that this work will serve as a useful foundation for the community to create new algorithms and make progress in the field of embodied AI multi-agent systems.
\vspace{-3mm}
 

\section{\simulator~Simulator}\label{sec:simulator}
Conventional multi-agent open-world simulators suffer from the gap between virtual agents and large-scale real-world humans. To bring this gap, we propose \simulator~with three key features: large-scale scalability, limited multimodal senses, and physical needs.
This section describes the design and implementation of this simulator, focusing on the architecture, observation space, state space, action space and communication. During introducing the design, we will highlight techniques that bring the three key features and omit the parts similar to other Minecraft simulators.

\begin{table}[t]
\small
\centering

\caption{Part of the comparison with related popular platforms or projects. The full table is in Table~\ref{table:full_comparison_of_platforms}. \textit{Max Agents}: The approximate maximum number of agents supported on a single PC (See Section \ref{subsec:expt_performance} for details).
\textit{Human}: Whether the simulator allows humans to directly interact with AI agents.
\textit{Plan as Action}: Whether agents can generate plans in a specific format (e.g., code) that the simulator interprets and executes directly as actions.
\textit{Sociological Experiments}: Whether the simulator is designed to facilitate the study of social phenomena and emergent behavior in multi-agent systems.
}
\label{table:comparison_of_platforms}
\renewcommand{\arraystretch}{1.1} 
\setlength\tabcolsep{4pt} 
\setlength{\tabcolsep}{3.2px}{
\begin{tabular}{lccccr}
\toprule
\multirow{2}{*}{\textbf{Simulator}} & \multirow{2}{*}{\textbf{Max Agents}} & \multirow{2}{*}{\textbf{Human}} & \multirow{2}{*}{\textbf{Plan as Action}} & \textbf{Sociological} & \multirow{2}{*}{\textbf{Number of Tasks}} \\
 & & & & \textbf{Experiments} & \\
\midrule
\textbf{\simulatorRaw} & 64+ & \checkmark & \checkmark & \checkmark & 6000+ \\
\midrule
MineDojo \cite{fan2022minedojo} & 1 & - & - & - & 3000+ \\
MineRL v1.0 \cite{baker2022video} & 1 & - & - & - & 5 \\
MarLÖ \cite{perezliebana2019multiagent} & 8 & \checkmark & - & - & 14 \\
Malmo \cite{johnson2016malmo}  & 8 & \checkmark & - & - & - \\
Voyager \cite{wang2023voyager} & 1 & \checkmark & \checkmark & - & - \\
Smallville \cite{park2023generative} & 25 & - & \checkmark & \checkmark & - \\
\bottomrule
\end{tabular}}
\renewcommand{\arraystretch}{1.0} 
\vspace{-3mm}
\end{table}

\subsection{Architecture}\label{sec:simulator_architecture}

\begin{figure}[t!]
    \centering
    \includegraphics[width=0.8\linewidth]{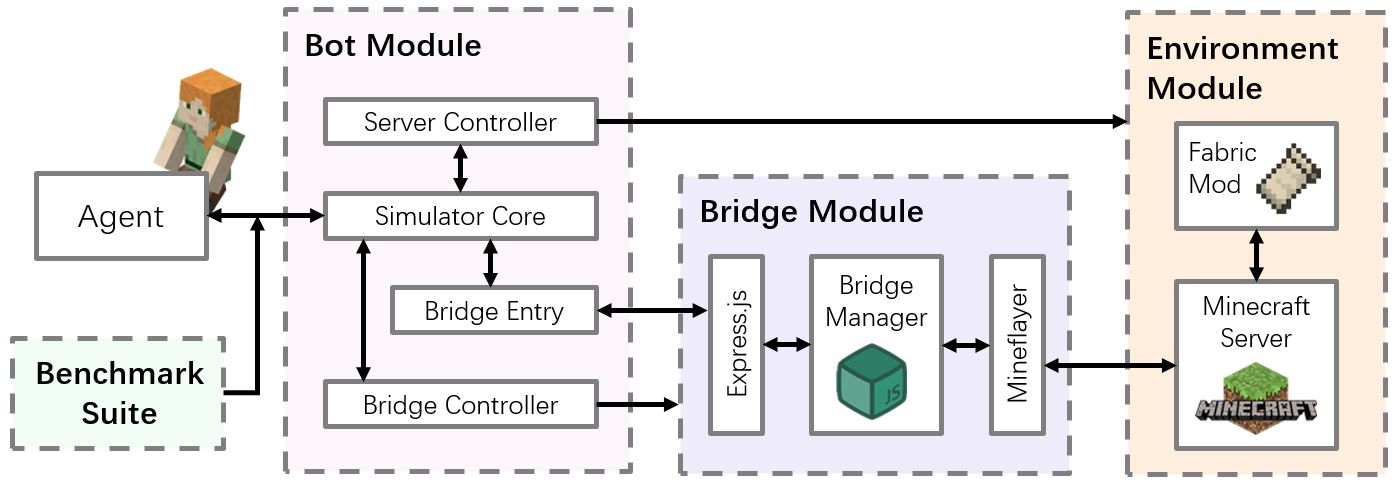}
    \caption{Illustration of the architecture of \simulator.}
    \label{fig:simulator}
    \vspace{-5mm}
\end{figure}

\simulator, inspired by Malmo \cite{johnson2016malmo}, Mineflayer~\cite{mineflayer}, MineDojo \cite{fan2022minedojo} and Voyager \cite{wang2023voyager}, is a Minecraft simulator where players\footnote{In this work, we use the term ``player'' to refer to both human players and AI agents. When we mention ``agent'', we specifically mean AI agents. Humans have the option to access the game either through VR or using a keyboard.} can explore, and interact with each other as well as the environments. 
The architecture as shown in Figure \ref{fig:simulator} consists of three main modules (Details in Section~\ref{sec:simulator_architecture_appendix}): the Bot Module, the Environment Module, and the Bridge Module.

\paragraph{What technology has caused the feature of large-scale scalability?}
Existing Minecraft simulators (e.g., Malmo, MineRL, MineDojo) necessitate running a Minecraft game client for each player. This client-based approach comes with a notable drawback - it incurs substantial resource costs and most machines cannot handle running large-scale Minecraft game clients concurrently. In contrast, \simulator~adopts a different approach. It simplifies each Minecraft client into a single thread, optimizing performance overhead caused by multiple clients. With \simulator, adding one more agent only requires one more thread.
Based on this new technique, \simulator~supports 64 or more agents on a mainstream consumer desktop PC, which is a substantial improvement compared to other Minecraft platforms that can only support up to 2 agents. We conducted relevant experiments in Section \ref{subsec:expt_performance}.


\subsection{Observation Space}
The observation space is designed to be compatible with almost all APIs of the popular MineDojo framework. \simulator~provides sensor information for players: tactile information (information about the blocks surrounding the agent, which represent the objects that the agent can touch), auditory information, and visual information (RGB video from the first-person perspective of the agent). These three modalities (namely, touch, vision, and hearing) together provide the agents with multimodal senses. Note that this information is all raw perceptual information.\footnote{Besides the raw perceptual information, \simulator~also provides the events encountered by the agent, such as injury, death, and others. Injury events can also be regarded as tactile information, but they are presented in the form of events for simplicity.}


\paragraph{What technology has caused the feature of limited multimodal senses?}
We refer to the mechanisms of human vision and hearing, and impose limitations on the sensor perception of players, including distance attenuation, environmental obstructions, and directional constraints, to model the limited senses.

\subsection{State Space}

Previous simulators focus on task-oriented activities, thus the state space focuses on the inventory and equipment. We use a state space that blends task-oriented activities with the rhythms of daily life. 

\paragraph{What technology has caused the feature of physical needs?}
Basic physical needs are the foundation that leads to daily life behavior. For the rhythms of daily life, we define the states of agents for themselves: physical needs like oxygen and hunger. 
Blending the rhythms of daily life with task-oriented activities is what makes this simulator stand out. Imagine agents waking up in their virtual Minecraft homes, engaging in daily routines like cooking to satisfy physical food needs, but also having defined jobs (e.g., lumberjack, farmer) that involve specific task-oriented activities. This creates a natural flow between daily life and goal-driven behavior, providing a more realistic and nuanced environment for studying agent interactions and complex social dynamics.

\subsection{Action Space}
The simulator offers a unique action space encompassing both low-level and high-level actions. For the low-level actions, \simulator~includes basic actions like walking, running, jumping, and interacting with objects. The low-level actions are the same as the traditional action space in gym-style API.
We also support Reinforcement Learning based on low-level actions (For more details about experiments of Reinforcement Learning method, please refer to Section \ref{subsec:expt_rl_agent}.
High-level actions, like dodging obstacles and manipulating tools, are suitable for complex tasks that consist of several low-level actions and require longer computation times. The high-level actions are implemented in the form of code, following Voyager \cite{wang2023voyager}. Imagine agents navigating the world, dodging obstacles, and manipulating tools. These complex tasks generate an action sequence, allowing the simulator to continue executing the low-level actions, skip some steps earlier, or be interrupted by some special event. 

\begin{figure*}[t!]
    \centering
    \includegraphics[width=\linewidth]{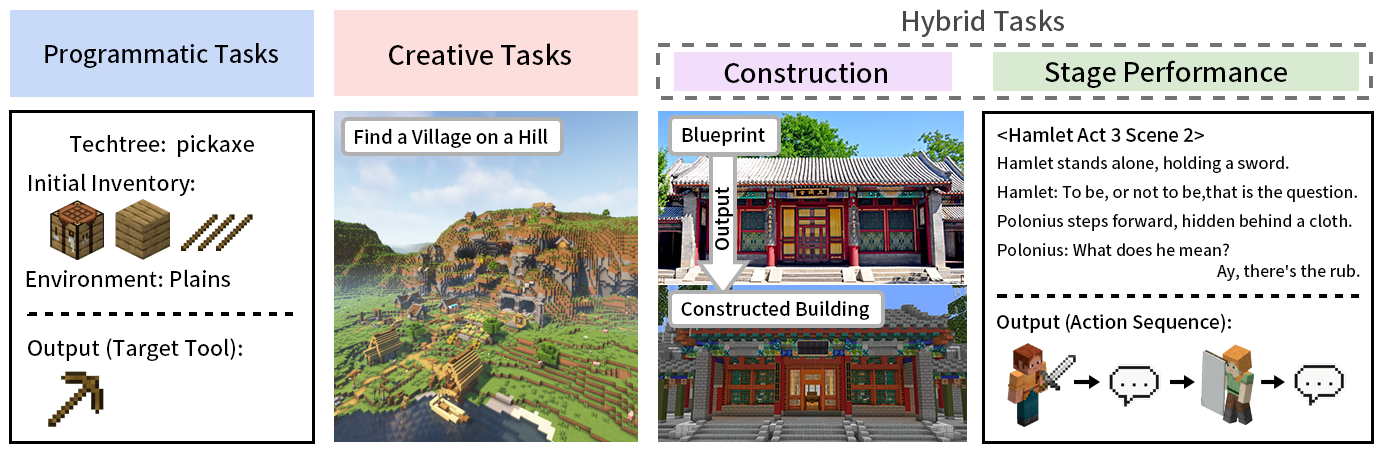}
    \caption{Illustration of Tasks. We have expanded the number of programmatic tasks and creative tasks by 2 times, compared to MineDojo. Additionally, we have introduced novel hybrid tasks that combine the features of programmatic tasks and creative tasks. Customizing the number of players is supported. For multi-agents, we provide two modes: cooperative mode and competitive mode.}
    \label{fig:task}
    \vspace{-5mm}
\end{figure*}

\subsection{Communication}

\paragraph{Diversity} Diverse communication strategies can lead to the emergence of more realistic behaviors within the simulated environment. We design three communication strategies including auditory information, body language (via visual perception), and sharing information in text media.

\paragraph{What communication technology has caused the feature of large-scale scalability?}
For multi-agent simulators, an efficient communication mechanism is crucial, especially in situations with large-scale scalability. Traditional communication methods, like centralized broadcasting to all agents, can become computationally expensive and slow down the simulation with a large number of agents. Here we design a distance-constrained constrained communication mechanism. If an agent wants to communicate with other agents, it chats through Minecraft's message bar. Only when the distance between other agents and the sending agent is less than a certain threshold, will other agents receive messages. Similarly, communications via auditory information and body language are limited by distance.

\paragraph{How is the interrupt mechanism implemented?}
Most importantly, the new message is allowed to interrupt the executing code and execute this message directly before the previous code has ended. We term this as the interrupt mechanism. With this mechanism, even if an agent is working on a 5-minute extension (such as mining), it is still feasible for other agents to communicate with this working agent at any time. This interrupt mechanism was not supported in the previous work, such as Voyager \cite{wang2023voyager}. Next, we will introduce how to implement it.
For a high-level action, the execution of the code is divided into several steps, with each step lasting 50-200 milliseconds\footnote{50 milliseconds is the minimum time unit in Minecraft. We refer to this minimum time unit as a ``tick''.}.
Before taking a step, the agent is provided with the running states of the previous code, either \textit{running}, \textit{ready}, or \textit{exceptions}. After completing a step, the agent, based on the running states, can choose to either switch to a new action code or continue executing the previous code. This function of choosing is implemented by an automatic gate control system with two gates: \textit{New} and \textit{Resume}. \textit{New} means the agent wants to switch to a new code in the following steps. \textit{Resume} indicates that the agent wants to continue executing the previous code. In this way, the agent can complete a code that needs to be executed for a long period, or be interrupted at an appropriate time. You may refer Section~\ref{subsec:multitasking_mechanism} for an example.

\section{\simulator~Benchmark Suite and Dataset}\label{sec:benchmark}



\simulator, as a large-scale multi-agent simulator, push the boundaries of multi-agent capabilities by enabling them to tackle complex human-like planning tasks within diverse environments. However, evaluating these advanced planning abilities necessitates sophisticated benchmarks. To address this challenge, we propose a new benchmark suit, \simulator~benchmark. \simulator~benchmark surpasses existing benchmarks by offering significantly more tasks (doubling programmatic and creative tasks compared to MineDojo) and introducing a novel ``hybrid task'' category that combines the features of programmatic and creative tasks (including Construction Tasks and Stage Performance Tasks). 
Additionally, this benchmark allows for flexible player numbers and exploration through cooperative and competitive modes. Competitive mode can be used to measure the differences in capabilities between different AI agents, as well as to develop adversarial learning algorithms.
Refer to Section~\ref{sec:appendix_benchmark_full_version} for more details.

\subsection{Programmatic Task}

We follow MineDojo \cite{fan2022minedojo} for the design of programmatic tasks.
Each Task $T$ is defined as a 5-tuple: $T=(G, \mathcal{G}, \mathcal{I}, f_{suc}, S)$. $G$ refers to the task goal that needs to be completed. $\mathcal{G}$ is guidance. $\mathcal{I}$ is the initial condition of the task. $f_{suc}$ is the Success Criterion. $S$ is a set of parameters that could be customized. Different from MineDojo, these parameters include the number of agents, cooperative mode, competitive mode, etc.
In total, \simulator~has 4499 programmatic tasks.


\subsection{Creative Task}
Creative tasks is defined by a 4-tuple: $T = (G, \mathcal{G}, \mathcal{I}, S)$.
There are 1536 creative tasks in total, and is compatible with tasks in MineDojo.

\subsection{Hybrid Task}
Hybrid tasks do not have a unique ground truth but have several references\footnote{Hybrid tasks resemble translation tasks in that, while a single unique translation may not exist, multiple reference translations can guide the process. Here references could be key rules, constraints, and key evaluation indicators, etc.}. We represent the hybrid task as 
$T = (G, \mathcal{G}, \mathcal{I}, \mathcal{D}, f_{score}, S)$. where $\mathcal{D}$ denotes the references. 
Unlike programmatic tasks, because Hybrid tasks do not have a ground truth, \simulator~will return a score of $f_{score}$ based on $\mathcal{D}$. The higher the score, the better the task is completed.
We design two types of tasks: Construction and Stage Performance, for hybrid tasks.

\textbf{Construction Tasks}
Given a blueprint for a building or scene, the Construction Task aims to build these buildings or scenes based on the blueprint. The blueprint, either pictures in real life or Minecraft-style pictures, is the reference $\mathcal{D}$.

\textit{Evaluation Metrics.} \simulator~gives a score based on whether the constructed buildings meet the blueprint's expectations.
We calculate the task scores through the VLM-based evaluations and human evaluation, which both use the same criteria (refer to Section~\ref{sec:construction-evaluation} for more details). The evaluation scores range from 1 to 5, with higher scores indicating that the agents' constructions more closely resemble the intended blueprint.

\textbf{Stage Performance Tasks}
Given a script of a drama consisting of several behaviors, which may be a single action or an emotional expression, the Stage Performance Task aims to perform the script with agents as actors. 

\textit{Evaluation Metrics.} 
We leverage two complementary evaluation metrics: an LCS (Longest Common Subsequence)-based metric and human evaluation.
The LCS-based metric consists of two distinct scores: the keypoint score and the appropriateness score. We present the formulas for adding up these two scores:
\begin{equation}
    f_{key} + f_{appro} = \frac{|LCS(SEQ_{Agent}, SEQ^{*})|}{|SEQ^{*}|} + \frac{|LCS(SEQ_{Agent}, SEQ^{*})|}{|SEQ_{Agent}|}
\end{equation}
where $SEQ_{Agent}$ represents the action sequence generated by the agent, while $SEQ^{*}$ is the ground truth. $LCS(A, B)$ denotes the LCS between $A$ and $B$. $|A|$ is the length of sequence $A$. 
The keypoint score emphasizes the completeness of the enacted behaviors, ensuring all crucial actions and expressions are performed. The appropriateness score goes beyond completeness to evaluate the overall coherence and naturalness of the performance, considering how well the behaviors flow together and align with the script's dramatic intent.
The human evaluation score is an integer between 1 and 5, with higher scores indicating better performance. It provides a comprehensive assessment of the agents' ability to not only execute actions accurately but also to deliver them naturally and engagingly. Details of the human evaluation criteria can be found in Section~\ref{sec:stage-performance-evaluation}.

\begin{figure*}[ht]
    \centering
    \includegraphics[width=0.9\linewidth]{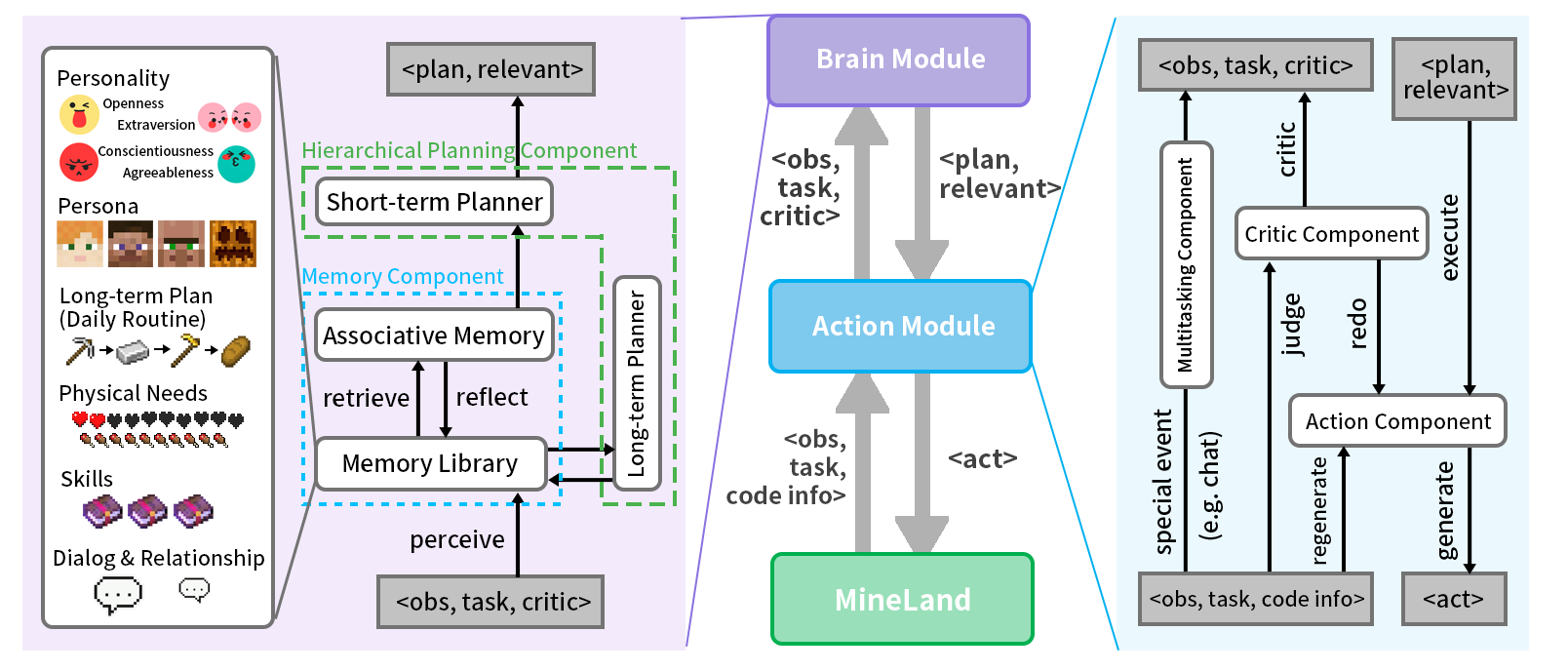}
    \caption{Illustration of the architecture of \agent.}
    \label{fig:agent}
    \vspace{-5mm}
\end{figure*}

\section{\agent~Agent}\label{sec:agent}
To truly demonstrate the challenge of this benchmark, we propose \agent\footnote{\agent~is the protagonist in the sandbox game Minecraft, one of the default skins for players and a character in the game: \url{https://www.minecraft.net/zh-hans}. To pay tribute, we named our proposed AI agent \agent.}, a VLM-based agent as shown in Figure~\ref{fig:agent}. Conventional LLM-powered AI agents depend on LLM to operate as its brain, which is backed by several vital components that perform various essential functions. These components, such as the memory component, planning component, and acting component, have been thoroughly studied recently. To cater to our specific requirements, we have improved these three components (refer to Section~\ref{sec:agent-full-version} for more details), replaced LLM with VLM and introduced one new component: the multitasking component. Additionally, \agent~exhibits different personality traits predefined in the system prompt.


\paragraph{Multitasking Component}\label{subsec:multitasking}
People often switch attention between tasks, for example cooking while talking.
The ability to communicate smoothly with other players while working on a task-oriented action is crucial in multi-agent scenarios. Therefore, we develop the mechanism of multitasking ability to enhance the agent's attention control and working memory abilities inspired by the Multitasking theory from the Cognition field~\cite{salvucci2008threaded}. 
Specifically, for attention control, the interrupt mechanism effectively controls attention among multiple tasks. For working memory, \agent~maintains and processes information in the Memory Library. When another agent says hello to the working agent, this involves saving and restoring internal states when frequent and high-speed switching between communication activities and goal-driven working actions to avoid forgetting ongoing tasks.
With the multitasking mechanism, \agent~allows for simultaneously simulating and executing intricate coordination and scheduling with multiple tasks.




\section{Experiments}\label{sec:exp}

\subsection{Experiments of Simulators Performance}\label{subsec:expt_performance}

We evaluate the number of agents that \simulator~can support and compare \simulator~with other popular Minecraft simulators. We utilize a mainstream consumer desktop PC equipped with an Intel i5-12400F CPU and 64GB of memory. Performance Monitor is employed to monitor the process. Our findings reveal that \simulator~is capable of supporting 32 agents simultaneously while providing visual displays. When visual display is disabled, the number of concurrently running agents increases to $\times$4 times. Furthermore, as depicted in Table~\ref{table:part-performance}, when \simulator~and Malmo both run 8 agents, \simulator's CPU and memory usage are approximately 1/3 that of Malmo's (specifically, 35.6\% and 38.0\%, respectively). It is worth noting that Malmo serves as the foundation for most popular Minecraft Platforms (e.g., MineDojo/MineRL/MarLÖ~\cite{perezliebana2019multiagent}), thus highlighting \simulator's superior performance compared to the vast majority of existing Minecraft Platforms. Consequently, \simulator~proves to be highly suitable for multi-agent environments.


\subsection{Experiments of Social Dynamics}\label{subsec:expt_social_dynamics}



In the ``unlocking tools'' task with two agents (Table \ref{multiagent_experiment}), two agents in the collaborative mode work together effectively, with a reduced workload per agent and higher communication expenses. On the other hand, two competitive agents worked independently and necessitated fewer code iterations. The primary reason is that, in competitive relationships, agents tend to achieve more in a single iteration to expedite progress and outperform their opponents. However, this results in less thorough planning and more code errors. Consequently, multiple agents in competitive relationships require fewer code iterations but make more mistakes.

We also observe that personality plays a significant role in determining the behavior of agents in multi-agent societies \cite{jiang2024evaluating}.
We assigned the personality traits of high extraversion and  agreeableness to both agents. Under this condition, the agents tended to establish collaboration and engage in mutual communication (co-op >8 out of 10 times). This result is consistent with human behavior \cite{de2000big}. When no personality was set for the agents, they worked independently (co-op 0 out of 10 times). For more experiments of simulating sociological phenomena with >10 agents, please refer to Section \ref{sec:exp_simulating_sociological_phenomena}.


\begin{table}[ht]
\centering
\begin{minipage}[t]{0.51\textwidth}
  \centering
  \small
  \caption{Part of the comparison table of performance of Minecraft simulators. The full table is in Table~\ref{table:full-performance}. \simulator$_{w/o\ vision}$ means \simulator~without vision. CPU and Mem represent the average CPU time and memory usage during the initialization phase and 5-minute run respectively.}
  \label{table:part-performance}
  \begin{tabular}{lccr}
    \toprule
    \textbf{Simulator} & \textbf{Agents} & \textbf{CPU} & \textbf{Mem} \\
    \midrule
    \simulatorRaw                 & 8  & 2.81\% & 7.07GB  \\
    \simulatorRaw                 & 32  & 5.64\% & 19.65GB  \\
    \simulatorRaw$_{w/o\ vision}$ & 8  & 1.88\% & 2.94GB  \\
    \simulatorRaw$_{w/o\ vision}$ & 64 & 2.87\% & 6.30GB  \\
    \simulatorRaw$_{w/o\ vision}$ & 128 & 4.00\% & 9.04GB  \\
    Malmo                         & 8  & 7.90\% & 18.63GB \\
    \bottomrule
  \end{tabular}
  \vspace{-4mm}
\end{minipage}%
\hfill
\begin{minipage}[t]{0.45\textwidth}
  \centering
  \small
  \caption{The number of code iterations needed per agent to unlock tools made of various materials is determined under three conditions. These conditions include a single agent, two agents in cooperation, and two agents in competition. Each experiment is repeated three times, and the success rate is 100\%.}
  \label{multiagent_experiment}
  \begin{tabular}{lccc}
    \toprule
    \multirow{2}{*}{\textbf{Relationship}} & \multicolumn{3}{c}{\textbf{Material}}\\ 
    \cmidrule{2-4}
    & Wooden & Stone & Iron \\
    \midrule
    Single Agent            & 7±2 & 10±3 & 25±7 \\
    Cooperative             & 13±5 & 20±7 & 49±10 \\
    Competitive             & 6±2 & 10±3 & 27±10  \\
    \bottomrule
  \end{tabular}
  \vspace{-4mm}
\end{minipage}
\end{table}

\subsection{Experiments of Construction Tasks}\label{subsec:exp_construction_task}
To establish a baseline for the research community, we evaluate our un-finetuned \agent~on two example
 construction tasks: ``Monument Construction'' and ``Stone Stele Construction'' (detailed in Section \ref{subsec:exp_construction_task-appendix} and Table \ref{construction_task_experiment}). While \agent~demonstrates a promising capability in selecting appropriate materials for both tasks, its overall construction abilities are currently limited. We posit two key reasons for these limitations: complex task planning and high-precision manipulation. For the first reason, construction tasks are inherently time-consuming and require sophisticated planning abilities. For instance, building auxiliary structures to facilitate the main construction process is crucial for achieving significant height. However, \agent~cannot currently plan and construct such auxiliary structures, hindering its ability to construct tall structures. The second reason is the high-precision manipulation. The construction tasks necessitate high-precision manipulation, such as adding decorative details. \agent~primarily utilizes high-level actions and lacks the necessary APIs to execute these fine-grained manipulations effectively. These results show the challenge of this task.


\subsection{Experiments of Stage Performance Tasks}\label{subsec:expt_stage_performance}

We evaluate our un-finetuned \agent~on four example stage performance tasks (Table \ref{stage_performance_experiment}). We found that in single-agent tasks, the agent typically exhibits high scores across all three evaluation metrics. However, in multi-agent tasks, the keypoint score tends to be higher than the appropriate score because the agent can understand the script and fulfill the key requirements. In contrast, the human evaluation scores are usually relatively lower because agents often engage in unnecessary dialogue in scripts that require collaboration, resulting in lower scores.


\begin{table}[ht]
\centering
\begin{minipage}[t]{0.35\textwidth}
  \centering
  \small
  \caption{Human evaluation score and VLM-based (gpt-4o) evaluation score of \agent~agent on two construction tasks. We conducted three evaluations using the VLM and obtained consistent results. Details in Section~\ref{sec:construction-evaluation}. VLM: VLM-based score. Human: Human evaluation score.}
  \label{construction_task_experiment}
  \begin{tabular}{lcc}
    \toprule
    \multirow{2}{*}{\textbf{Construction}} & \multicolumn{2}{c}{\textbf{Score}}\\ 
    \cmidrule{2-3}
    & VLM & Human \\
    \midrule
    A Monument            & 4, 3, 3 & 3 \\
    A Stone Stele           & 1, 1, 1 & 1 \\
    \bottomrule
  \end{tabular}
\end{minipage}%
\hfill
\begin{minipage}[t]{0.62\textwidth}
  \centering
  \small
  \caption{Appropriateness score (Appro.), keypoint score (Key.), and Human Evaluation Score (Human.) of \agent~Agent across four stage performance tasks. The number in parentheses following the script name represents the number of agents in that script. Appropriateness score and keypoint score are real numbers in the range \( [0, 1] \), and the human evaluation score is an integer between 1 and 5. Details in Section~\ref{sec:exp_stage_performance_tasks_appendix} and Section~\ref{sec:stage-performance-evaluation}.}
  \label{stage_performance_experiment}
  \begin{tabular}{lccr}
    \toprule
    \multirow{2}{*}{\textbf{Script Name}} & \multicolumn{3}{c}{\textbf{Score}}\\ 
    \cmidrule{2-4}
    & Appro. & Key. & Human.\\
    \midrule
    Cook food(1)             & 1.00 & 1.00 & 5 \\
    Exchange items(2)           & 0.59 & 0.99 & 4 \\
    Make friends(3)            & 0.67 & 0.98 & 3 \\
    Romeo and Julia, Act I Scene I(13) & 0.09 & 0.20 & 1 \\
    \bottomrule
  \end{tabular}
\end{minipage}
\vspace{-5mm}
\end{table}

\subsection{Experiments of Multi-Agent Cooperation}\label{subsec:expt_multi_agent}
To validate the cooperation efficiency of our agent framework, we conduct the ``unlocking tools'' task with two agents. We observed that agents in a cooperative relationship required more code iterations to finish the task, primarily because most of these iterations were dedicated to establishing and maintaining communication, as well as task allocation. For example, when one agent says in a chat that he needs two sticks, another agent will ask for getting together near the  table, and then give the sticks to him. However, the actual workload for each agent is reduced without considering the chat cost. Compared to agents working independently, the code iteration cost of agents cooperating is reduced by 20\% per agent.

\subsection{Experiments of Multitasking}\label{subsec:expt_multitasking}

To validate the impact of multitasking support in the simulator\footnote{This multitasking functionality is achieved through the synergistic interplay of two keys: the multitasking component in \agent~and the interrupt mechanism in \simulator.}, we conduct the obsidian mining task, which takes over 8 minutes and requires multiple steps to complete. In the beginning, the agent is mining and encounters chat or hurt events within 8 minutes. 
\textbf{Chat event}: Another agent nearby initiates a conversation with \agent. This is a low-priority event, and \agent~can choose whether to respond to the agent nearby. \textbf{Hurt event}: The agent gets hurts. For example, a zombie attacks the agent. This is a high-priority event, requiring \agent~to stop its current task and address this event first.

We expect the two types of events to activate the multitasking component and the agent processes the mining and the special event simultaneously.
Results are in Table \ref{comparison_of_agent_multitasking}. 
In all ten runs, we count the number of successfully handling multitasks. For the hurt event, if the agent fights back due to this hurt event, it is considered successful. For the chat event, if the agent responds due to the chat event, it is considered successful.
Results reveal that \agent$_{w/o\ mt}$ can't process events timely, resulting in \agent~being killed by the zombie. In contrast, \agent~with a multitasking component is capable of managing multiple events (e.g., mining while chatting), autonomously determining their priority, and addressing the higher-priority events first. Hence, multitasking is an essential mechanism.


\begin{table}[ht]
\small
\centering
\caption{Comparison between \agent~and \agent$_{w/o\ mt}$ (\agent~without the multitasking component).}
\label{comparison_of_agent_multitasking}
\begin{tabular}{lcc}
\toprule
\textbf{Agent} & \textbf{Hurt event} & \textbf{Chat event}\\
\midrule
\agentRaw~                & 8/10 & 2/10 \\ 
\agentRaw$_{w/o\ mt}$ & 0/10 & 0/10 \\
\bottomrule
\end{tabular}
\vspace{-4mm}
\end{table}


\section{Conclusion}\label{sec:conclusion}
Traditional multi-agent simulators struggle with large-scale scenarios and often assume perfect information and unrealistic agent capabilities. To address these limitations, we introduce \simulator, a novel Minecraft-based simulator supporting 64 or more agents with limited senses and physical needs. This forces agents to actively communicate and collaborate, fostering more ecologically valid interactions. The advantage carries potential broader impacts across various domains as discussed in Section~\ref{sec:broader_impact}.

{
    \bibliographystyle{plain}
    \bibliography{paper}

\begin{thebibliography}{10}

\bibitem{alderfer1969empirical}
Clayton~P Alderfer.
\newblock An empirical test of a new theory of human needs.
\newblock {\em Organizational behavior and human performance}, 4(2):142--175,
  1969.

\bibitem{bai2023there}
Jitao Bai, Simiao Zhang, and Zhonghao Chen.
\newblock Is there any social principle for llm-based agents?
\newblock {\em arXiv preprint arXiv:2308.11136}, 2023.

\bibitem{baker2022video}
Bowen Baker, Ilge Akkaya, Peter Zhokhov, Joost Huizinga, Jie Tang, Adrien
  Ecoffet, Brandon Houghton, Raul Sampedro, and Jeff Clune.
\newblock Video pretraining (vpt): Learning to act by watching unlabeled online
  videos, 2022.

\bibitem{22_bates1994role}
Joseph Bates.
\newblock The role of emotion in believable agents.
\newblock {\em Communications of the ACM}, 37(7):122--125, 1994.

\bibitem{binz2023using}
Marcel Binz and Eric Schulz.
\newblock Using cognitive psychology to understand gpt-3.
\newblock {\em Proceedings of the National Academy of Sciences},
  120(6):e2218523120, 2023.

\bibitem{41_Bledsoe1986Dream}
Woody Bledsoe.
\newblock I had a dream: Aaai presidential address.
\newblock {\em AI Magazine}, 7(1):57--61, 1986.

\bibitem{cai2023open}
Shaofei Cai, Zihao Wang, Xiaojian Ma, Anji Liu, and Yitao Liang.
\newblock Open-world multi-task control through goal-aware representation
  learning and adaptive horizon prediction.
\newblock {\em arXiv preprint arXiv:2301.10034}, 2023.

\bibitem{Cai_2023_CVPR}
Shaofei Cai, Zihao Wang, Xiaojian Ma, Anji Liu, and Yitao Liang.
\newblock Open-world multi-task control through goal-aware representation
  learning and adaptive horizon prediction.
\newblock In {\em Proceedings of the IEEE/CVF Conference on Computer Vision and
  Pattern Recognition (CVPR)}, pages 13734--13744, June 2023.

\bibitem{cao2012overview}
Yongcan Cao, Wenwu Yu, Wei Ren, and Guanrong Chen.
\newblock An overview of recent progress in the study of distributed
  multi-agent coordination.
\newblock {\em IEEE Transactions on Industrial informatics}, 9(1):427--438,
  2012.

\bibitem{card1983psychology}
Stuart~K Card, Thomas~P Moran, and Alan Newell.
\newblock The psychology of human-computer interaction.
\newblock 1983.

\bibitem{chen2024sagent}
Jiaqi Chen, Yuxian Jiang, Jiachen Lu, and Li~Zhang.
\newblock S-agent: self-organizing agents in open-ended environment.
\newblock In {\em ICLR 2024 Workshop on Large Language Model (LLM) Agents},
  2024.

\bibitem{da2019variational}
Ant{\^o}nio~Carlos da~Rocha~Costa.
\newblock {\em A Variational Basis for the Regulation and Structuration
  Mechanisms of Agent Societies}.
\newblock Springer, 2019.

\bibitem{de2000big}
Boele De~Raad.
\newblock {\em The big five personality factors: the psycholexical approach to
  personality.}
\newblock Hogrefe \& Huber Publishers, 2000.

\bibitem{29_dill2011game}
Kevin Dill and L~Martin.
\newblock A game ai approach to autonomous control of virtual characters.
\newblock In {\em Proceedings of the Interservice/Industry Training,
  Simulation, and Education Conference (I/ITSEC'11)}, Orlando, FL, USA, 2011.

\bibitem{doyal1984theory}
Len Doyal and Ian Gough.
\newblock A theory of human needs.
\newblock {\em Critical Social Policy}, 4(10):6--38, 1984.

\bibitem{fan2022minedojo}
Linxi Fan, Guanzhi Wang, Yunfan Jiang, Ajay Mandlekar, Yuncong Yang, Haoyi Zhu,
  Andrew Tang, De-An Huang, Yuke Zhu, and Anima Anandkumar.
\newblock Minedojo: Building open-ended embodied agents with internet-scale
  knowledge.
\newblock In {\em Thirty-sixth Conference on Neural Information Processing
  Systems Datasets and Benchmarks Track}, 2022.

\bibitem{feng2023llama}
Yicheng Feng, Yuxuan Wang, Jiazheng Liu, Sipeng Zheng, and Zongqing Lu.
\newblock Llama rider: Spurring large language models to explore the open
  world, 2023.

\bibitem{gong2023mindagent}
Ran Gong, Qiuyuan Huang, Xiaojian Ma, Hoi Vo, Zane Durante, Yusuke Noda, Zilong
  Zheng, Song-Chun Zhu, Demetri Terzopoulos, Li~Fei-Fei, et~al.
\newblock Mindagent: Emergent gaming interaction.
\newblock {\em arXiv preprint arXiv:2309.09971}, 2023.

\bibitem{gussminerl}
William~H Guss, Brandon Houghton, Nicholay Topin, Phillip Wang, Cayden Codel,
  Manuela Veloso, and Ruslan Salakhutdinov.
\newblock Minerl: A large-scale dataset of minecraft demonstrations.
\newblock {\em arXiv preprint arXiv:1907.13440}, 2019.

\bibitem{guss2019minerl}
William~H. Guss, Brandon Houghton, Nicholay Topin, Phillip Wang, Cayden Codel,
  Manuela Veloso, and Ruslan Salakhutdinov.
\newblock Minerl: A large-scale dataset of minecraft demonstrations, 2019.

\bibitem{heil1983perception}
John Heil.
\newblock Perception and cognition.
\newblock 1983.

\bibitem{20_hollan1984steamer}
James~D. Hollan, Edwin~L. Hutchins, and Louis Weitzman.
\newblock Steamer: An interactive inspectable simulation-based training system.
\newblock {\em AI Magazine}, 5(2):23--36, 1984.

\bibitem{63_horton2023large}
John~J. Horton.
\newblock Large language models as simulated economic agents: What can we learn
  from homo silicus?, 2023.

\bibitem{jiang2023evaluating}
Guangyuan Jiang, Manjie Xu, Song-Chun Zhu, Wenjuan Han, Chi Zhang, and Yixin
  Zhu.
\newblock Evaluating and inducing personality in pre-trained language models.
\newblock In {\em Thirty-seventh Conference on Neural Information Processing
  Systems}, 2023.

\bibitem{jiang2024evaluating}
Guangyuan Jiang, Manjie Xu, Song-Chun Zhu, Wenjuan Han, Chi Zhang, and Yixin
  Zhu.
\newblock Evaluating and inducing personality in pre-trained language models.
\newblock {\em Advances in Neural Information Processing Systems}, 36, 2024.

\bibitem{john1996goms}
Bonnie~E John and David~E Kieras.
\newblock The goms family of user interface analysis techniques: Comparison and
  contrast.
\newblock {\em ACM Transactions on Computer-Human Interaction (TOCHI)},
  3(4):320--351, 1996.

\bibitem{johnson2016malmo}
M.~Johnson, K.~Hofmann, T.~Hutton, and D.~Bignell.
\newblock The malmo platform for artificial intelligence experimentation.
\newblock In {\em Proc. 25th International Joint Conference on Artificial
  Intelligence}, page 4246, Palo Alto, California USA, 2016. AAAI Press.

\bibitem{11_jones1999automated}
Randolph~M Jones, John~E Laird, Paul~E Nielsen, Karen~J Coulter, Patrick Kenny,
  and Frank~V Koss.
\newblock Automated intelligent pilots for combat flight simulation.
\newblock {\em AI Magazine}, 20(1):27--42, 1999.

\bibitem{juslin2005vocal}
Patrik~N Juslin, Klaus~R Scherer, J~Harrigan, and R~Rosenthal.
\newblock Vocal expression of affect.
\newblock {\em The new handbook of methods in nonverbal behavior research},
  pages 65--135, 2005.

\bibitem{kiseleva2021neurips}
Julia Kiseleva, Ziming Li, Mohammad Aliannejadi, Shrestha Mohanty, Maartje ter
  Hoeve, Mikhail Burtsev, Alexey Skrynnik, Artem Zholus, Aleksandr Panov, Kavya
  Srinet, Arthur Szlam, Yuxuan Sun, Katja Hofmann, Michel Galley, and Ahmed
  Awadallah.
\newblock Neurips 2021 competition iglu: Interactive grounded language
  understanding in a collaborative environment, 2021.

\bibitem{ai2thor}
Eric Kolve, Roozbeh Mottaghi, Winson Han, Eli VanderBilt, Luca Weihs, Alvaro
  Herrasti, Daniel Gordon, Yuke Zhu, Abhinav Gupta, and Ali Farhadi.
\newblock {AI2-THOR: An Interactive 3D Environment for Visual AI}.
\newblock {\em arXiv}, 2017.

\bibitem{4_laird2001human}
John Laird and Michael VanLent.
\newblock Human-level ai's killer application: Interactive computer games.
\newblock {\em AI Magazine}, 22(2):15, 2001.

\bibitem{lakkanige2023exploring}
Kalyani Lakkanige, Lamar Cooley-Russ, Alan~R. Wagner, and Sarah Rajtmajer.
\newblock Exploring trust and risk during online bartering interactions, 2023.

\bibitem{leibo2021meltingpot}
Joel~Z. Leibo, Edgar~Du\'e\ nez Guzm\'an, Alexander~Sasha Vezhnevets, John~P.
  Agapiou, Peter Sunehag, Raphael Koster, Jayd Matyas, Charles Beattie, Igor
  Mordatch, and Thore Graepel.
\newblock Scalable evaluation of multi-agent reinforcement learning with
  melting pot.
\newblock PMLR, 2021.

\bibitem{li2022igibson}
Chengshu Li, Fei Xia, Roberto Mart\'in-Mart\'in, Michael Lingelbach, Sanjana
  Srivastava, Bokui Shen, Kent~Elliott Vainio, Cem Gokmen, Gokul Dharan, Tanish
  Jain, Andrey Kurenkov, Karen Liu, Hyowon Gweon, Jiajun Wu, Li~Fei-Fei, and
  Silvio Savarese.
\newblock igibson 2.0: Object-centric simulation for robot learning of everyday
  household tasks.
\newblock In Aleksandra Faust, David Hsu, and Gerhard Neumann, editors, {\em
  Proceedings of the 5th Conference on Robot Learning}, volume 164 of {\em
  Proceedings of Machine Learning Research}, pages 455--465. PMLR, 08--11 Nov
  2022.

\bibitem{li2024auto}
Hao Li, Xue Yang, Zhaokai Wang, Xizhou Zhu, Jie Zhou, Yu~Qiao, Xiaogang Wang,
  Hongsheng Li, Lewei Lu, and Jifeng Dai.
\newblock Auto mc-reward: Automated dense reward design with large language
  models for minecraft, 2024.

\bibitem{liu2024rlgpt}
Shaoteng Liu, Haoqi Yuan, Minda Hu, Yanwei Li, Yukang Chen, Shu Liu, Zongqing
  Lu, and Jiaya Jia.
\newblock Rl-gpt: Integrating reinforcement learning and code-as-policy, 2024.

\bibitem{madge2024large}
Chris Madge and Massimo Poesio.
\newblock Large language models as minecraft agents, 2024.

\bibitem{openai2024gpt4}
OpenAI.
\newblock Gpt-4 technical report, 2024.

\bibitem{park2023generative}
Joon~Sung Park, Joseph O'Brien, Carrie~Jun Cai, Meredith~Ringel Morris, Percy
  Liang, and Michael~S Bernstein.
\newblock Generative agents: Interactive simulacra of human behavior.
\newblock In {\em Proceedings of the 36th Annual ACM Symposium on User
  Interface Software and Technology}, pages 1--22, 2023.

\bibitem{9_park2022socialsimulacra}
Joon~Sung Park, Lindsay Popowski, Carrie~J. Cai, Meredith~Ringel Morris, Percy
  Liang, and Michael~S. Bernstein.
\newblock Social simulacra: Creating populated prototypes for social computing
  systems.
\newblock In {\em In the 35th Annual ACM Symposium on User Interface Software
  and Technology (UIST '22)}, UIST '22, New York, NY, USA, 2022. Association
  for Computing Machinery.

\bibitem{perezliebana2019multiagent}
Diego Perez-Liebana, Katja Hofmann, Sharada~Prasanna Mohanty, Noburu Kuno,
  Andre Kramer, Sam Devlin, Raluca~D. Gaina, and Daniel Ionita.
\newblock The multi-agent reinforcement learning in malm\"o (marl\"o)
  competition, 2019.

\bibitem{mineflayer}
PrismarineJS.
\newblock mineflayer.
\newblock \url{https://github.com/PrismarineJS/mineflayer}, 2023.

\bibitem{puig2018virtualhome}
Xavier Puig, Kevin Ra, Marko Boben, Jiaman Li, Tingwu Wang, Sanja Fidler, and
  Antonio Torralba.
\newblock Virtualhome: Simulating household activities via programs.
\newblock In {\em Proceedings of the IEEE Conference on Computer Vision and
  Pattern Recognition}, pages 8494--8502, 2018.

\bibitem{puig2020watch}
Xavier Puig, Tianmin Shu, Shuang Li, Zilin Wang, Yuan-Hong Liao, Joshua~B
  Tenenbaum, Sanja Fidler, and Antonio Torralba.
\newblock Watch-and-help: A challenge for social perception and human-ai
  collaboration.
\newblock In {\em International Conference on Learning Representations}, 2020.

\bibitem{puig2020watchandhelp}
Xavier Puig, Tianmin Shu, Shuang Li, Zilin Wang, Joshua~B. Tenenbaum, Sanja
  Fidler, and Antonio Torralba.
\newblock Watch-and-help: A challenge for social perception and human-ai
  collaboration, 2020.

\bibitem{puig2023habitat}
Xavier Puig, Eric Undersander, Andrew Szot, Mikael~Dallaire Cote, Tsung-Yen
  Yang, Ruslan Partsey, Ruta Desai, Alexander~William Clegg, Michal Hlavac,
  So~Yeon Min, et~al.
\newblock Habitat 3.0: A co-habitat for humans, avatars and robots.
\newblock {\em arXiv preprint arXiv:2310.13724}, 2023.

\bibitem{qian2023communicative}
Chen Qian, Xin Cong, Cheng Yang, Weize Chen, Yusheng Su, Juyuan Xu, Zhiyuan
  Liu, and Maosong Sun.
\newblock Communicative agents for software development.
\newblock {\em arXiv preprint arXiv:2307.07924}, 2023.

\bibitem{qin2024mp5}
Yiran Qin, Enshen Zhou, Qichang Liu, Zhenfei Yin, Lu~Sheng, Ruimao Zhang,
  Yu~Qiao, and Jing Shao.
\newblock Mp5: A multi-modal open-ended embodied system in minecraft via active
  perception, 2024.

\bibitem{radford2021learning}
Alec Radford, Jong~Wook Kim, Chris Hallacy, Aditya Ramesh, Gabriel Goh,
  Sandhini Agarwal, Girish Sastry, Amanda Askell, Pamela Mishkin, Jack Clark,
  et~al.
\newblock Learning transferable visual models from natural language
  supervision.
\newblock In {\em International conference on machine learning}, pages
  8748--8763. PMLR, 2021.

\bibitem{stable-baselines3}
Antonin Raffin, Ashley Hill, Adam Gleave, Anssi Kanervisto, Maximilian
  Ernestus, and Noah Dormann.
\newblock Stable-baselines3: Reliable reinforcement learning implementations.
\newblock {\em Journal of Machine Learning Research}, 22(268):1--8, 2021.

\bibitem{6_riedl2012interactive}
Mark~O. Riedl.
\newblock Interactive narrative: A novel application of artificial intelligence
  for computer games.
\newblock In {\em Proceedings of the Twenty-Sixth AAAI Conference on Artificial
  Intelligence (AAAI'12)}, pages 2160--2165, 2012.

\bibitem{salvucci2008threaded}
Dario~D Salvucci and Niels~A Taatgen.
\newblock Threaded cognition: an integrated theory of concurrent multitasking.
\newblock {\em Psychological review}, 115(1):101, 2008.

\bibitem{shen2021igibson}
Bokui Shen, Fei Xia, Chengshu Li, Roberto Mart\'in-Mart\'in, Linxi Fan, Guanzhi
  Wang, Claudia P\'erez-D'Arpino, Shyamal Buch, Sanjana Srivastava, Lyne~P.
  Tchapmi, Micael~E. Tchapmi, Kent Vainio, Josiah Wong, Li~Fei-Fei, and Silvio
  Savarese.
\newblock igibson 1.0: a simulation environment for interactive tasks in large
  realistic scenes.
\newblock In {\em 2021 IEEE/RSJ International Conference on Intelligent Robots
  and Systems (IROS)}, page accepted. IEEE, 2021.

\bibitem{szot2021habitat}
Andrew Szot, Alex Clegg, Eric Undersander, Erik Wijmans, Yili Zhao, John
  Turner, Noah Maestre, Mustafa Mukadam, Devendra Chaplot, Oleksandr Maksymets,
  Aaron Gokaslan, Vladimir Vondrus, Sameer Dharur, Franziska Meier, Wojciech
  Galuba, Angel Chang, Zsolt Kira, Vladlen Koltun, Jitendra Malik, Manolis
  Savva, and Dhruv Batra.
\newblock Habitat 2.0: Training home assistants to rearrange their habitat.
\newblock In {\em Advances in Neural Information Processing Systems (NeurIPS)},
  2021.

\bibitem{10_tambe1995intelligent}
Milind Tambe, W~Lewis Johnson, Randolph~M Jones, Frank Koss, John~E Laird,
  Paul~S Rosenbloom, and Karl Schwamb.
\newblock Intelligent agents for interactive simulation environments.
\newblock {\em AI Magazine}, 16(1):15, 1995.

\bibitem{wang2023voyager}
Guanzhi Wang, Yuqi Xie, Yunfan Jiang, Ajay Mandlekar, Chaowei Xiao, Yuke Zhu,
  Linxi Fan, and Anima Anandkumar.
\newblock Voyager: An open-ended embodied agent with large language models.
\newblock {\em arXiv preprint arXiv:2305.16291}, 2023.

\bibitem{wang2023jarvis1}
Zihao Wang, Shaofei Cai, Anji Liu, Yonggang Jin, Jinbing Hou, Bowei Zhang,
  Haowei Lin, Zhaofeng He, Zilong Zheng, Yaodong Yang, Xiaojian Ma, and Yitao
  Liang.
\newblock Jarvis-1: Open-world multi-task agents with memory-augmented
  multimodal language models.
\newblock {\em arXiv preprint arXiv: 2311.05997}, 2023.

\bibitem{wang2023describe}
Zihao Wang, Shaofei Cai, Anji Liu, Xiaojian Ma, and Yitao Liang.
\newblock Describe, explain, plan and select: Interactive planning with large
  language models enables open-world multi-task agents.
\newblock {\em arXiv preprint arXiv:2302.01560}, 2023.

\bibitem{williams2023epidemic}
Ross Williams, Niyousha Hosseinichimeh, Aritra Majumdar, and Navid
  Ghaffarzadegan.
\newblock Epidemic modeling with generative agents.
\newblock {\em arXiv preprint arXiv:2307.04986}, 2023.

\bibitem{wimmer2021everyday}
S~Wimmer, A~Pfeiffer, and N~Denk.
\newblock The everyday life in the sims 4 during a pandemic. a life simulation
  as a virtual mirror of society?
\newblock In {\em INTED2021 Proceedings}, pages 5754--5760. IATED, 2021.

\bibitem{Xiang_2020_SAPIEN}
Fanbo Xiang, Yuzhe Qin, Kaichun Mo, Yikuan Xia, Hao Zhu, Fangchen Liu, Minghua
  Liu, Hanxiao Jiang, Yifu Yuan, He~Wang, Li~Yi, Angel~X. Chang, Leonidas~J.
  Guibas, and Hao Su.
\newblock {SAPIEN}: A simulated part-based interactive environment.
\newblock In {\em The IEEE Conference on Computer Vision and Pattern
  Recognition (CVPR)}, June 2020.

\bibitem{yu2023kola}
Jifan Yu, Xiaozhi Wang, Shangqing Tu, Shulin Cao, Daniel Zhang-Li, Xin Lv, Hao
  Peng, Zijun Yao, Xiaohan Zhang, Hanming Li, et~al.
\newblock Kola: Carefully benchmarking world knowledge of large language
  models.
\newblock {\em arXiv preprint arXiv:2306.09296}, 2023.

\bibitem{zhang2023creative}
Chi Zhang, Penglin Cai, Yuhui Fu, Haoqi Yuan, and Zongqing Lu.
\newblock Creative agents: Empowering agents with imagination for creative
  tasks, 2023.

\bibitem{zhang2023building}
Hongxin Zhang, Weihua Du, Jiaming Shan, Qinhong Zhou, Yilun Du, Joshua
  Tenenbaum, Tianmin Shu, and Chuang Gan.
\newblock Building cooperative embodied agents modularly with large language
  models.
\newblock In {\em NeurIPS 2023 Foundation Models for Decision Making Workshop},
  2023.

\bibitem{zhao2023think}
Zhonghan Zhao, Wenhao Chai, Xuan Wang, Li~Boyi, Shengyu Hao, Shidong Cao, Tian
  Ye, Jenq-Neng Hwang, and Gaoang Wang.
\newblock See and think: Embodied agent in virtual environment, 2023.

\bibitem{zhao2024hierarchical}
Zhonghan Zhao, Kewei Chen, Dongxu Guo, Wenhao Chai, Tian Ye, Yanting Zhang, and
  Gaoang Wang.
\newblock Hierarchical auto-organizing system for open-ended multi-agent
  navigation, 2024.

\bibitem{zhou2024minedreamer}
Enshen Zhou, Yiran Qin, Zhenfei Yin, Yuzhou Huang, Ruimao Zhang, Lu~Sheng,
  Yu~Qiao, and Jing Shao.
\newblock Minedreamer: Learning to follow instructions via chain-of-imagination
  for simulated-world control, 2024.

\bibitem{zhu2023ghost}
Xizhou Zhu, Yuntao Chen, Hao Tian, Chenxin Tao, Weijie Su, Chenyu Yang, Gao
  Huang, Bin Li, Lewei Lu, Xiaogang Wang, Yu~Qiao, Zhaoxiang Zhang, and Jifeng
  Dai.
\newblock Ghost in the minecraft: Generally capable agents for open-world
  environments via large language models with text-based knowledge and memory.
\newblock {\em arXiv preprint arXiv:2305.17144}, 2023.

\bibitem{zhuge2023mindstorms}
Mingchen Zhuge, Haozhe Liu, Francesco Faccio, Dylan~R Ashley, R{\'o}bert
  Csord{\'a}s, Anand Gopalakrishnan, Abdullah Hamdi, Hasan Abed Al~Kader
  Hammoud, Vincent Herrmann, Kazuki Irie, et~al.
\newblock Mindstorms in natural language-based societies of mind.
\newblock {\em arXiv preprint arXiv:2305.17066}, 2023.

\end{thebibliography}
}


\newpage

\appendix

\section{Related Work}\label{sec:related_work}
\subsection{Multi-Agent Simulator}
As the popularity of AI agent research continues to grow, there has also been a focus on studying multiple AI agents as well as their cooperation and competition. Researchers and practitioners imagine a dynamic artificial society where human interactions can be simulated by trustworthy agents~\cite{da2019variational}. From two individuals~\cite{cao2012overview,puig2020watch}, through four individuals~\cite{wimmer2021everyday}, to sandbox games Smallville with twenty-five individuals~\cite{park2023generative}, we witness how individuals perceive a simulated society as the backdrop and interact with the agents and people who engage with it. Each individual can be portrayed through a program, a real human, or an agent based on LLM~\cite{park2023generative}. The interaction between individuals plays a role in shaping social behavior, leading to simulation of the society. Simulating larger societies can be advantageous. Increasing the number of agents can lead to greater specialization, enabling the accomplishment of more complex and larger-scale tasks. This can significantly improve task efficiency, such as in software development tasks~\cite{qian2023communicative}.
Additionally, such simulation of interaction has had a significant effect in many other fields. For example, it can replicate realistic social phenomena~\cite{29_dill2011game, 9_park2022socialsimulacra}, enhance social robots~\cite{22_bates1994role, 41_Bledsoe1986Dream}. They can also be used to test social science theories~\cite{binz2023using,jiang2023evaluating,63_horton2023large}, create model human processors for theory and usability testing~\cite{card1983psychology, john1996goms}, train people on how to handle rare yet difficult interpersonal situations~\cite{10_tambe1995intelligent, 11_jones1999automated, 20_hollan1984steamer}, and support game characters~\cite{4_laird2001human, 6_riedl2012interactive}.
There are also many non-Minecraft and non-open-world simulators, such as Habitat3.0 \cite{puig2023habitat}, VirtualHome \cite{puig2018virtualhome, puig2020watchandhelp}, SAPIEN \cite{Xiang_2020_SAPIEN}, AI2-THOR \cite{ai2thor}, and iGibson \cite{li2022igibson, shen2021igibson}.

\textbf{Challenges of Scaling Up the Number of Agents.}
While increasing the number of agents can improve task efficiency and make multi-agent simulations more realistic~\cite{qian2023communicative,park2023generative,williams2023epidemic}, current research primarily focuses on a small number of agents~\cite{park2023generative,bai2023there,zhuge2023mindstorms}. This is mainly due to the challenges of scaling up the number of agents. Deploying a large number of AI agents will result in an increased computational burden, necessitating better architectural design and computational optimization~\cite{park2023generative}. Most research in terms of AI agents mimicking daily life routines, focused on two agents~\cite{puig2020watch}. Simulators that remind people of sandbox games (The Sims) initially support four individuals~\cite{wimmer2021everyday}, then are extended to twenty-five individuals by \cite{park2023generative}.

\textbf{Challenges of Limited Multimodal Senses.}
To ensure the authenticity of the simulation, an ideal multi-agent simulator should operate under the fundamental assumption that agents possess only limited multimodal senses like humans~\cite{heil1983perception}. Limited multimodal senses mean partially observable environments and an eco-centric perspective.
Limited visual and auditory senses restrict information access, forcing agents to actively navigate and communicate to compensate for sensory deficiencies. This mirrors real-life social interactions, where visibility and audibility can be affected by factors such as distance, terrain, and context~\cite{juslin2005vocal}. As the number of agents grows, the challenges of limited multimodal senses become quite difficult. This is because the communication network of the entire system becomes highly intricate.
For agents in our \simulator, the video input is an eco-centric perspective (first perspective) instead of the third perspective in \cite{park2023generative}, which is omniscient and unrealistic.

\textbf{Challenges of Physical Needs.}
Multi-agent simulation platforms hold immense potential for exploring and understanding human social dynamics. However, existing paradigms often disregard the human needs~\cite{doyal1984theory,alderfer1969empirical}. Existing simulators are designed to simulate believable human behavior in daily-life activities~\cite{park2023generative,bai2023there}. These activities include waking up, cooking breakfast, heading to work, and initiating conversations with others. However, they do not define their physical needs. For example, after a certain amount of time has passed, the agent will become hungry and have the desire to cook. This desire then leads to the next actions. In this way, the action of cooking is motivated by real desires instead of a predefined schedule.

We incorporate practical physical requirements into the agent model. Agents have basic physical needs: sleep, food, and resource management, which introduces an engaging time-based element to their daily routine processes. This encourages collaboration and competition for resources, reflecting the intricate balance of cooperation and self-interest seen in human societies.

\subsection{Multi-Agent Simulator w.r.t Minecraft}
Minecraft, the beloved sandbox game, has been a valuable platform for researchers exploring various fields, including artificial intelligence and multi-agent systems, because of its open world and diverse mechanics.

Specifically, the flexibility and richness of Minecraft make it perfect for developing multi-agent simulators. Researchers have the ability to create various custom environments and scenarios within the game world, where they can introduce virtual agents with specific goals and capabilities. These agents can then interact with each other and the environment, providing researchers with the opportunity to observe and analyze their behavior in a controlled setting. These simulators can be broadly categorized into two main types: task-oriented simulators and daily-life simulators.

Task-oriented simulators focus on agents achieving specific objectives within a set time frame. Minecraft is regarded as the training ground for AI agents to hone their skills. For example, \cite{gussminerl,fan2022minedojo, zhou2024minedreamer, liu2024rlgpt} focuses on exploring the environment and gathering resources like wood and stone, and managing them efficiently to complete tasks like building structures or crafting tools. Agents in \cite{zhao2024hierarchical} are designed to complete navigation tasks. In \cite{chen2024sagent}, agents are self-organized to execute collaborative building tasks and resource collection tasks. Agents in \cite{gong2023mindagent} must work together to overcome challenges that require joint effort.

Daily-life simulators take a more holistic approach, focusing on the daily lives of agents within a virtual society. \cite{park2023generative} simulates the rhythms and routines of daily life. This includes activities such as waking up, cooking breakfast, going to work, forming opinions, observing others, and engaging in conversations. \cite{lakkanige2023exploring} simulates sociological experiments, and examines how risk affects the way people engage in bartering.

\subsection{AI Agent with LLMs and VLMs}
LLMs or VLMs are commonly utilized to bootstrap the components of the Agent. In particular, LLMs have demonstrated effective performance for task-planning~\cite{gong2023mindagent,zhu2023ghost,madge2024large,li2024auto,feng2023llama}, and they possess substantial world knowledge~\cite{yu2023kola}. Team CraftJarvis has also developed a powerful agent, \cite{wang2023jarvis1}, based on their previous projects \cite{wang2023describe,cai2023open}. Moreover, VLMs like CLIP~\cite{radford2021learning} offer a versatile visual-language representation that aligns with language and enables zero-shot visual recognition capabilities for potential AI agents. Using VLMs in construction tasks has many advantages over existing methods~\cite{zhang2023creative}. AI agents with visual information can also complete tasks in human-like ways~\cite{qin2024mp5,zhao2023think}. 

\section{Broader Impact}\label{sec:broader_impact}
By including the simulator, benchmark, and agent framework, we allow for research on more authentic and detailed interactions in simulated environments. 

\paragraph{Advancing AI Multi-Agent Research}
The proposed simulator, \simulator, offers a platform for studying agents interacting under more realistic conditions. This can lead to the development of more robust and adaptable AI agents capable of effectively collaborating and navigating complex social scenarios. These advancements hold immense potential for various applications, including human-computer interaction, robotics, and game design.

\paragraph{Understanding Human Dynamics}
By studying agent interactions within the simulator, users may gain valuable insights into human social dynamics. Analyzing collaboration, communication, and competition in this controlled environment can help us understand real-world social phenomena and predict their potential outcomes. 

\paragraph{Ethics Statement}
This study follows the ethical principles stated in the Declaration of Helsinki. All participants will receive comprehensive information about the nature and objectives of the study and will be required to provide written consent. Participation in this study is voluntary, and participants have the right to withdraw at any time without facing any consequences. The confidentiality and privacy of participants will be safeguarded in accordance with relevant laws and regulations.

\section{Details of \simulator~Simulator}
We show the comparison with related popular platforms or projects in Table~\ref{table:full_comparison_of_platforms}.

\begin{table}[h]
\small
\centering

\caption{Comparison with related popular platforms or projects. \textit{Max Agents}: The approximate maximum number of agents supported on a single PC (See Section \ref{subsec:expt_performance} for details).
\textit{Human}: Whether the simulator allows humans to directly interact with AI agents.
\textit{Plan as Action}: Whether agents can generate plans in a specific format (e.g., code) that the simulator interprets and executes directly as actions.
\textit{Sociological Experiments}: Whether the simulator is designed to facilitate the study of social phenomena and emergent behavior in multi-agent systems.
\textit{Physical Needs}: Whether agents have simulated physiological needs (e.g., hunger, thirst) that influence their behavior and require actions to fulfill. \textit{Open World}: Open-world or not. \textit{3D Space}: 3D environment or not. 
}
\label{table:full_comparison_of_platforms}
\renewcommand{\arraystretch}{1.1} 
\setlength\tabcolsep{4pt} 
\setlength{\tabcolsep}{3.2px}{
\begin{tabular}{lcccccccr}
\toprule
\multirow{2}{*}{\textbf{Simulator}} & \textbf{Max} & \multirow{2}{*}{\textbf{Human}} & \textbf{Plan as} & \textbf{Sociological} & \textbf{Physical} & \textbf{Open} & \textbf{3D} & \textbf{Number} \\
 & \textbf{Agents} & & \textbf{Action} & \textbf{Experiments} & \textbf{Needs} & \textbf{World} & \textbf{Space} & \textbf{of Tasks} \\
\midrule
\textbf{\simulatorRaw} & 64+ & \checkmark & \checkmark & \checkmark & \checkmark & \checkmark & \checkmark & 6000+ \\
\midrule
MineDojo \cite{fan2022minedojo} & 1 & - & - & - & \checkmark & \checkmark & \checkmark & 3000+ \\
MineRL v0.4 \cite{guss2019minerl} & 1 & - & - & - & \checkmark & \checkmark & \checkmark & 11 \\
MineRL v1.0 \cite{baker2022video} & 1 & - & - & - & \checkmark & \checkmark & \checkmark & 5 \\
MarLÖ \cite{perezliebana2019multiagent} & 8 & \checkmark & - & - & \checkmark & \checkmark & \checkmark & 14 \\
Malmo \cite{johnson2016malmo}  & 8 & \checkmark & - & - & \checkmark & \checkmark & \checkmark & - \\
IGLU \cite{kiseleva2021neurips} & 1 & - & - & - & \checkmark & \checkmark & \checkmark & 157 \\
Voyager \cite{wang2023voyager} & 1 & \checkmark & \checkmark & - & \checkmark & \checkmark & \checkmark & - \\
\midrule
Habitat 3.0 \cite{puig2023habitat} & 2+ & \checkmark & - & - & - & - & \checkmark & 200 \\
Habitat 2.0 \cite{szot2021habitat} & 1 & - & - & - & - & - & \checkmark & 105 \\
VirtualHome-Social \cite{puig2020watchandhelp} & 40+ & \checkmark & \checkmark & \checkmark & - & - & \checkmark & 50 \\
SAPIEN \cite{Xiang_2020_SAPIEN} & 1 & - & - & - & - & - & \checkmark & - \\
AI2-THOR \cite{ai2thor} & 6 & - & - & - & - & - & \checkmark & 200+ \\
iGibson \cite{li2022igibson} & 2+ & \checkmark & - & - & - & - & \checkmark & 50 \\
Melting Pot \cite{leibo2021meltingpot} & 50+ & \checkmark & - & \checkmark & - & - & - & 256 \\
Smallville \cite{park2023generative} & 25 & - & \checkmark & \checkmark & \checkmark & \checkmark & - & - \\
\bottomrule
\end{tabular}}
\renewcommand{\arraystretch}{1.0} 
\end{table}

\subsection{Architecture}\label{sec:simulator_architecture_appendix}

The architecture consists of three main modules: the Bot Module, the Environment Module, and the Bridge Module.

\noindent\textit{Bot Module:} Providing the Minecraft environment information to the agent and implementing a series of APIs that agents can use to control entities. \\
\noindent\textit{Environment Module:} Collecting the environment information, passing environment feedback to the bridge module, executing the action in the environment (by operating the Fabric server instance), and offering some APIs enabling the bot module to alter the server state.\\
\noindent\textit{Bridge Module:} Serving as a bridge, to transfer the environment information and agent-generated action. \footnote{Bridge Module is based on Mineflayer, which benefits from an excellent community (\url{https://github.com/PrismarineJS}). We have improved community tools.}
\subsection{State Space}\label{subsec:state-space-appendix}
We focus on a state space that blends task-oriented activities with the rhythms of daily life. This blending opens up exciting possibilities for studying dynamic and ecologically valid multi-agent interactions. For example, agents have to balance the goal of mining with the need to fill their stomachs, otherwise they will starve to death.
 Here are detailed explanations for the state space:
\begin{itemize}
    \item \textbf{Health}: Indicate the agent's current health status, which can be affected by sleep and enemy attacks. It is represented as an integer in the range [0, 20]. A higher value represents better health.
    \item \textbf{Food}: Indicate the agent's level of satiety. A higher value represents better satiety.
    \item \textbf{Oxygen}: When the agent sinks into water, an oxygen tank will appear and begin to consume oxygen. 
    \item \textbf{Inventory}: Represent all the resources owned by the agent in their backpack, like the potion in the backpack.
    \item \textbf{Equipment}: Indicate the equipment worn by the agent, like a sword in the hand.
\end{itemize}

\subsection{Action Space}\label{subsec:action-space-appendix}
In contrast to generating textual action descriptions and training an external controller module \cite{Cai_2023_CVPR} for transforming the plan to the executable code, this simulator is a language-model-friendly simulator, providing the code instead of the action description. 
Through the code, it can directly execute plans generated by the language model.
Agents in \simulator~generate a series of codes based on the Mineflayer API while representing planning, such as moving, watching, and mining. The advantage of using code is that it avoids error accumulation, whereas using a textual plan requires an additional model to map the plan to the code, which can lead to error accumulation. 

\subsection{Supporting up to 100 Agents}
\begin{figure*}[ht]
    \centering
    \includegraphics[width=\linewidth]{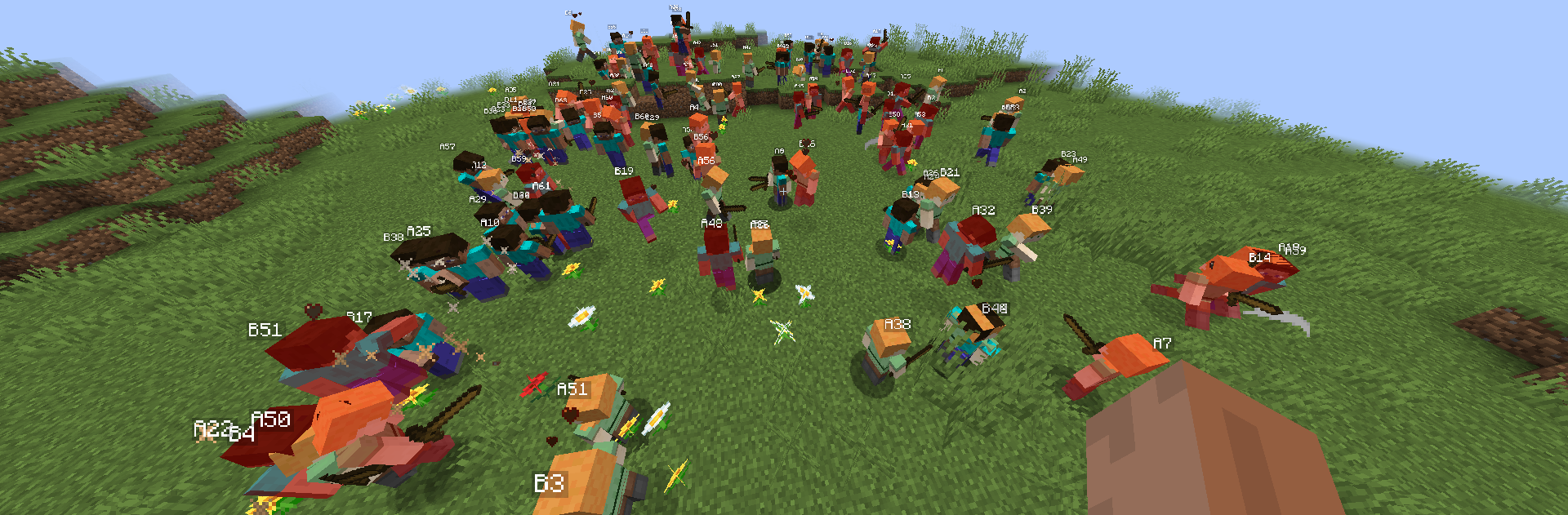}
    \caption{100 agents are fighting within the \simulator~Simulator.}
    \label{fig:100agents}
\end{figure*}

To evaluate the \simulator~simulator's ability to handle large-scale agents, we designed an experiment featuring 100 agents engaged in a simultaneous combat scenario (Figure \ref{fig:100agents}). This high agent count pushes the boundaries of previously typical two-agent simulations and demonstrates the simulator's scalability in effectively managing a massive number of agents acting concurrently.

\section{Details of \simulator~Benchmark Suite}\label{sec:appendix_benchmark_full_version}
\simulator~Benchmark Suite offers a wide and diverse range of tasks, including three categories: programmatic tasks, creative tasks, and hybrid tasks.
We referred to MineDojo for the design of programmatic tasks. Programmatic tasks are divided into four categories: Survival, Harvest, Tech Tree, and Combat. The Survival task requires the agent to survive for a specific number of days without dying. The Harvest task requires the agent to obtain certain specific items. The Tech Tree task requires the agent to obtain certain specific tools. The Combat task requires the agent to kill certain specific creatures or enemies.
Tech Tree Tasks require agents to make specific tools that represent the current level of Agent technology development. Combat tasks require agents to defeat certain creatures. Survival tasks require the agent to survive for a period of time. The metrics for these four tasks are the probability of success for multiple evaluation episodes, the number of in-game ticks, and the number of code iterations.
Creative Tasks will give the agent an open task objective to facilitate exploration.
We have formalized the definition of tasks, allowing developers to easily add new tasks.

\begin{figure*}[ht]
    \centering
    \includegraphics[width=\linewidth]{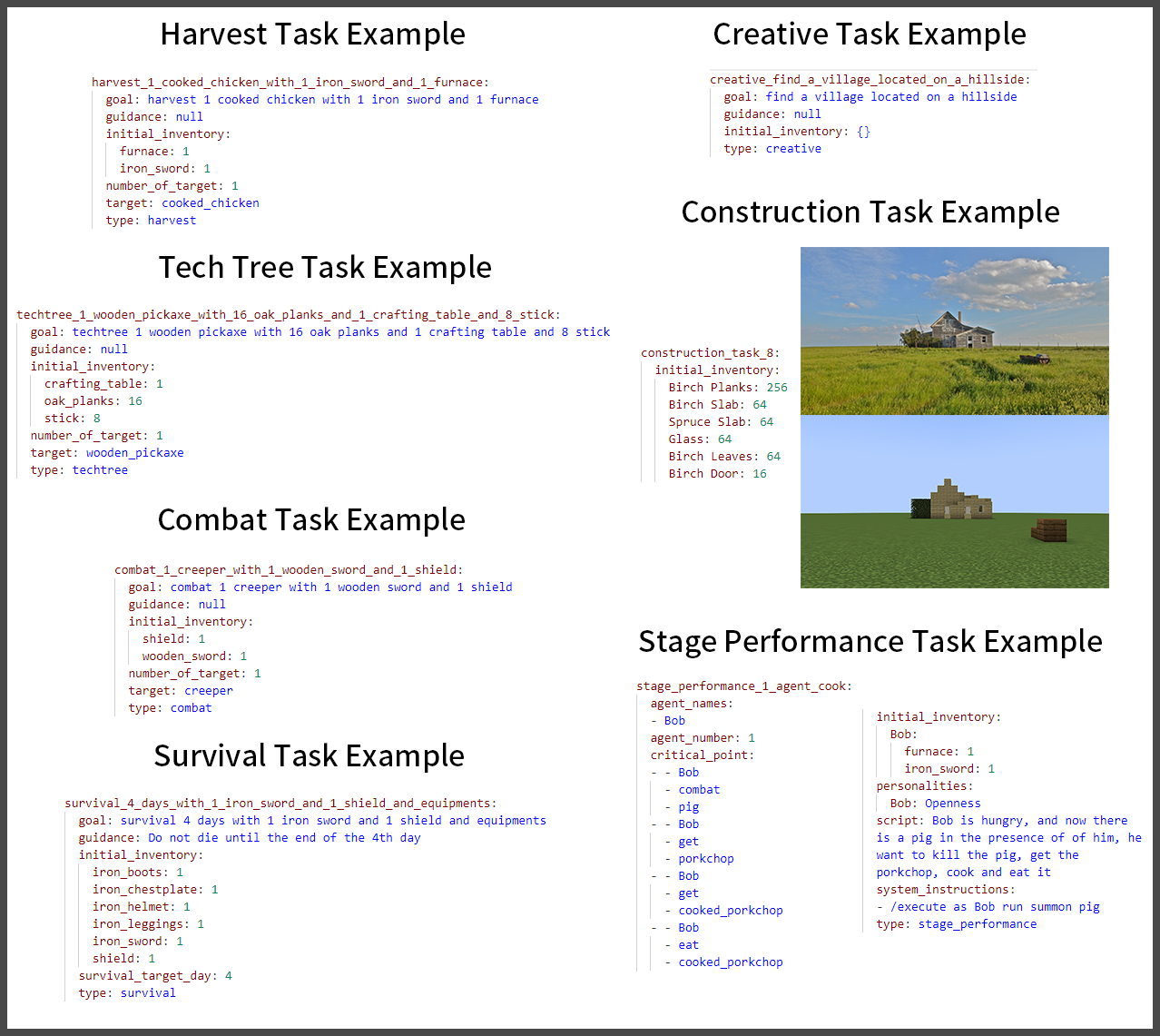}
    \caption{Illustration of cases for different tasks.}
    \label{fig:task_example}
\end{figure*}

\begin{figure}[ht]
    \centering
    \includegraphics[width=\linewidth]{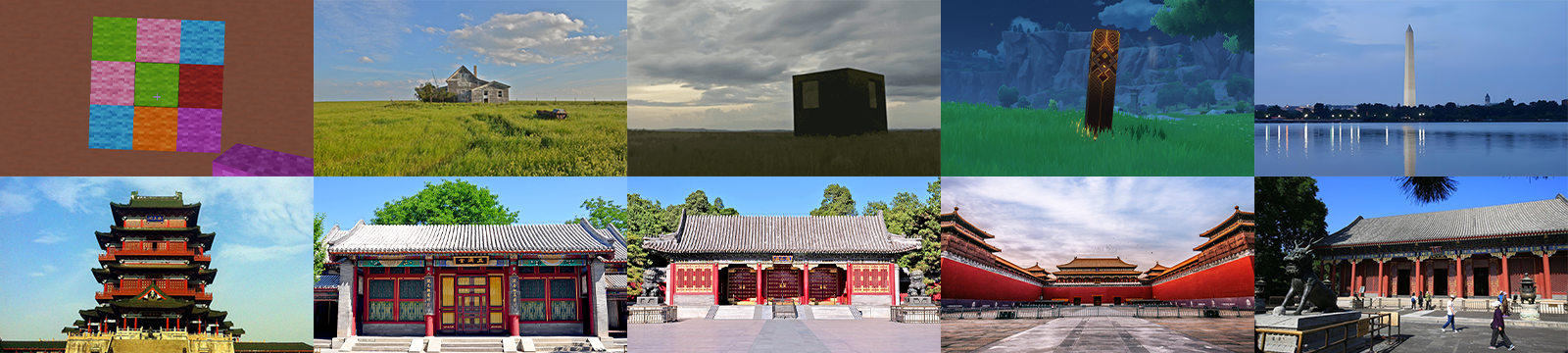}
    \caption{Some of the blueprints for the construction tasks.}
    \label{fig:blueprints- construction-tasks}
\end{figure}

The data statistics of the dataset are shown in Table~\ref{table:task_number}. Figure~\ref{fig:task_example} displays specific task data. We show several blueprints in Figure~\ref{fig:blueprints- construction-tasks}.

\begin{table}[ht]
\centering
\caption{Statistical analysis of the tasks in the \simulator~Benchmark Suite.}
\label{table:task_number}
\begin{tabular}{lr}
\toprule
\textbf{Task Category} & \textbf{Number of Tasks} \\
\midrule
Harvest Tasks   & 1361 \\
Tech Tree Tasks & 861 \\
Combat Tasks    & 2232 \\
Survival Tasks  & 45  \\
\midrule
Creative Tasks From \simulatorRaw & 12   \\
Creative Tasks From MineDojo & 1524 \\
\midrule
Construction Tasks      & 13 \\
Stage Performance Tasks & 5  \\
\midrule
All Tasks & 6053 \\
\bottomrule
\end{tabular}
\end{table}

\section{Details of \agent}\label{sec:agent-full-version}
Crafting an AI agent for \simulator, where daily life seamlessly blends with task-oriented activities, opens up exciting possibilities. The agent should fulfill daily needs like cooking, socializing, and maintaining shelter, while also completing assigned tasks like resource gathering, crafting, or construction. We propose \agent, a VLM-based approach, to balance daily routines and tasks. \agent~supports both individual daily-life goals and community-based task-oriented goals. 


Our agent design addresses the unique requirements of our large-scale multi-agent \simulator~simulator. Unlike prior simulators focused on smaller agent populations, our environment necessitates the incorporation of a social brain module. This module equips the agent with the ability to navigate complex and multitasking social interactions and make informed decisions within a crowded environment.

Furthermore, our Minecraft-inspired simulator introduces the challenge of action failure, unlike environments where ``action spoken equals success'' (referencing Smallville \cite{park2023generative}). To address this, we have designed an action correction module. This module functions similarly to human neural reflexes, handling minor action corrections and alleviating the burden on the social brain module. This division of labor allows the social brain to focus on higher-level decision-making while the action correction module ensures smoother execution.

Different from most conventional LLM-based agents, \agent~process information from various sources like visual, auditory (hearing conversations, environmental sounds), and tactile (touching objects) to build a comprehensive understanding of the world. For the remaining parts of \simulator~that are not emphasized, we used the default setting of Voyager \cite{wang2023voyager}.
AI agents perceive the environment, make plans, execute plans independently, and interact with other agents. Internally, \agent~may exhibit different personality traits predefined in the system prompt.
It utilizes the VLM for processing sensor information within the Minecraft environment. \agent~also tracks its own states, energy levels, and resource inventory to inform its actions and prioritize tasks.




\subsection{Memory Component}\label{subsec:memory}



The brain module can be considered as composed of the memory component and the planning component. The memory library is responsible for the storage and retrieval of memories, managing all memories in \agent's life. The planning component, based on memories and external information, generates a plan for the action module to execute.
In this subsection, we will detail the memory component, which consists of two main parts: memory library and associative memory.

The memory library is responsible for storing all of the agent's information and retrieving relevant information from memory based on events. The memory library stores the agent's personality, persona, long-term goals, short-term goals, chat records, experienced events, mastered skills, and environmental information. The memory library also includes a long-term planner. When accessing the memory library for the first time, the long-term planner generates a long-term plan based on the personality, persona, and observations, which is then stored in the memory library. In each iteration, the memory library processes information from observations, critic information, and tasks, and stores it in the vector database \footnote{ 
 We leverage Chroma in \agent~for storing and retrieving memory. \url{https://www.trychroma.com/}}. If the critic information indicates that the previous task has been completed, the skill manager in the memory library will be called upon to generate a concise description of the relevant skill, which will then be stored in the skill vector database.

Associative memory is responsible for storing \agent's short-term memories and relevant memories related to the current situation, aiding the short-term planner to focus on important rather than irrelevant information. Upon a special event, \agent~first decides in the associative memory whether the event requires high-priority processing; if so, it interrupts the current code to generate a new short-term plan.

There is a bidirectional communication mechanism between the memory library and associative memory. The memory library extracts relevant information according to the current situation and stores it in associative memory, which then returns the generated short-term plans to be stored in the memory library.

\subsection{Hierarchical Planning Component}\label{subsec:planner}
Based on observation, inner states, task running states, and event information, \simulator~considers the current task's complexity degree. If it is complex, it will generate a long-term plan for later decomposition into short-term plans; Otherwise, a short-term plan will be generated directly and executed immediately. Next, all the information related to the plan is integrated into the associative memory and memory library, including the generated long-term plan.
Afterward, \agent~extracts information from associative memory and generates short-term plans and explanations, which are then input into the action module.

Because of the multi-tiered goals of \simulator, we implement a hierarchical planning system with different levels of abstraction. The top-level (i.e., long-term planning) focuses on long-term community goals and individual aspirations, while the lower levels handle specific tasks and sub-goals within the daily schedule.
With long-term planning and short-term planning, agents can pursue bigger objectives within the daily life context, such as building a community, accumulating wealth, or achieving societal goals. This adds a layer of strategic planning and foresight to their behavior. 

Additionally, different from other task-oriented AI agents, the long-term planning module interleaves daily routines and tasks: design the planner to seamlessly interweave daily routines like cooking or socializing with task-oriented actions like resource gathering or construction. This ensures the agent fulfills both individual needs and community objectives efficiently.


\subsection{Action Module}\label{subsec:self_correction}



The action module is responsible for converting short-term plans along with related information into code, executing the code, and performing the self-correction circle. The action module includes three components:

\begin{itemize}
    \item \textbf{Action component}: Responsible for converting the plan into specific steps and codes.
    \item \textbf{Critic component}: Used to detect whether a certain execution result conforms to the short-term plan, so as to determine whether the current plan has been completed.
    \item \textbf{Dispatching component}: Responsible for receiving environmental information and distributing it to the other two components according to different situations.
\end{itemize}

This self-correction circle allows for identifying and correcting deviations from planned behaviors or task execution.
This equips the agent with mechanisms to detect and recover from errors like missed goals, failed actions, or unforeseen consequences. 
In previous work, such as Smallville \cite{park2023generative}, action execution will not fail. Planning to cook will definitely lead to success. However, in \simulator~as well as reality, actions may fail due to various reasons, such as unexpected events.
So, we need self-correction to solve some simple action errors in the action module. Inspired by Voyager \cite{wang2023voyager}, we implement self-correction. By reading information such as observation space and short-term plan, we comprehensively consider and determine the completion status of the short-term plan through the critic component. 

\subsection{Multitasking Mechanism}\label{subsec:multitasking_mechanism}
We'll explain the multitasking mechanism from both the simulator side and the agent side.
From the simulator-side event management, the simulator maintains a dedicated event queue that stores various special events categorized by their urgency (e.g., hurt events, chat events, and death events).
In terms of agents, when an agent encounters an event from the queue, it employs a prioritization mechanism to assess its importance and urgency. Lower-priority events, such as chat events, can be deferred to minimize disruption of the ongoing task. The agent's decision-making module (referred to as the brain module) plays a central role. If the event necessitates attention, the brain module can either:

\begin{itemize}
    \item \textbf{Context Switching}: The brain module temporarily stores the ongoing task in the memory library and focuses on addressing the new event. This facilitates context switching when necessary.
    \item \textbf{Concurrent Processing}: In specific scenarios, the brain module might choose to handle both tasks simultaneously, leveraging its multitasking capabilities.
\end{itemize}
This strategy, involving the simulator's event queue and the agent's prioritization and execution mechanisms, fosters effective multitasking within the multi-agent environment.

\section{Details of Hyper-Parameters}
\label{sec:appendix_hyper_params}
\agent~leverage OpenAI's \textit{gpt-4-vision-preview}\cite{openai2024gpt4} API with a temperature of 0 for text completion in all components, and \textit{text-embedding-ada-002} API for text embedding in the memory library for storage and retrieve memory. Apart from the action component, the maximum tokens are set to 512. For the action component, the maximum tokens are expanded to 512$\times$3.

\begin{itemize}
    \item \textbf{AI Agent}: \agent's personality and persona in our experiments is ``None'' in the default situation.
    \item \textbf{Dispatching component}: ``FAILED TIMES LIMIT'' refers to the maximum number of attempts allowed when \agent~has a code error. The default value is 3. ``code execution time limit'' is 2000 ticks.
    \item \textbf{Critic component}: ``FAILED TIMES LIMIT'' refers to the maximum number of attempts allowed when \agent~failed to achieve a short-term plan. The default value is 2. ``Critic Mode'' is ``auto'' when we allow agents to automatically judge whether to achieve the short-term plan.
    \item \textbf{Memory Library}: ``chat retrieve limit'' is 5. ``event retrieve limit'' is 2. ``environment retrieve limit'' is 2. ``skill retrieve limit'' is 5. ``recent chat retrieve limit'' is 8. ``short term plan retrieve limit'' is 5.
    \item \textbf{\simulatorRaw~\& Minecraft Server}: By default, ``gamemode'' is survival mode. ``difficulty'' is peaceful mode (normal in combat tasks). ``view-distance'' on server side is 6. ``simulation-distance'' is 3. ``task mode'' is cooperative mode.
\end{itemize}

\section{Experiments of Minecraft Simulator Performance Comparison}\label{sec:appendix_data}
We evaluate the number of agents that \simulator~can support and compare \simulator~with other popular Minecraft simulators, in Table~\ref{table:full-performance}. We utilize a mainstream consumer desktop PC equipped with an Intel i5-12400F CPU and 64GB of memory. 
Vision condition means \simulator~provides visual display and headless mode. The headless mode means \simulator~doesn't provide a visual display. The two values of the Vision column (e.g., 6 and 250ms) are view distance and visual refresh interval in milliseconds, respectively.

Our findings reveal that \simulator~is capable of supporting up to 32 agents simultaneously while providing a visual display, and up to 128 agents in the headless mode without visual displays. Furthermore, when \simulator~and Malmo both run 8 agents, \simulator's CPU and memory usage is approximately 1/3 that of Malmo's (specifically, 35.6\% and 38.0\%, respectively). It is worth noting that Malmo serves as the foundation for most popular Minecraft Platforms (e.g., MineDojo/MineRL v0.4/MarLÖ), thus highlighting \simulator's superior performance compared to the vast majority of existing Minecraft Platforms. 

It should be noted that original MineDojo and MineRL only support a single agent, while MarLÖ by default supports only 2 agents. Malmo originally supported only one agent.
We have designed a new task for Malmo to test the maximum number of agents it can handle. The result is 8.


\begin{table}[ht]
\centering
\small
\caption{Comparison of performance between \simulator~and other popular Minecraft simulators under different conditions. \simulator* represents an optimized version, so its Max CPU is significantly reduced.}
\label{table:full-performance}
\begin{tabular}{lccccr}
\toprule
\textbf{Simulator} & \textbf{Conditions} & \textbf{Agents}  & \textbf{Avg CPU} &\textbf{Max CPU}& \textbf{Avg Mem} \\
\midrule
\simulatorRaw & Vision = (6, 250ms) & 1  & 1.74\% & 23.22\% & 4.11 GB\\
\simulatorRaw & Vision = (6, 250ms) & 4  & 2.80\% & 48.16\% & 5.16 GB\\
\simulatorRaw & Vision = (6, 250ms) & 8  & 4.85\% & 63.60\% & 7.87 GB\\
\simulatorRaw & Vision = (3, 500ms) & 1  & 1.54\% & 18.92\% & 3.61 GB\\
\simulatorRaw & Vision = (3, 500ms) & 4  & 2.09\% & 31.00\% & 5.06 GB\\
\simulatorRaw & Vision = (3, 500ms) & 8  & 2.81\% & 33.08\% & 7.07 GB\\
\simulatorRaw & Vision = (3, 500ms) & 16 & 4.66\% & 46.63\% & 10.17 GB\\
\simulatorRaw*& Vision = (3, 500ms) & 32 & 5.64\% & 40.15\% & 19.65 GB\\
\simulatorRaw & Headless Mode       & 1  & 1.47\% & 19.91\% & 3.23 GB\\
\simulatorRaw & Headless Mode       & 4  & 1.73\% & 27.27\% & 3.30 GB\\
\simulatorRaw & Headless Mode       & 8  & 1.88\% & 26.83\% & 2.94 GB\\
\simulatorRaw & Headless Mode       & 16 & 2.72\% & 45.17\% & 3.51 GB\\
\simulatorRaw & Headless Mode       & 24 & 3.92\% & 81.64\% & 3.65 GB\\
\simulatorRaw & Headless Mode       & 32 & 2.81\% & 73.14\% & 4.64 GB\\
\simulatorRaw & Headless Mode       & 40 & 2.98\% & 84.56\% & 5.34 GB\\
\simulatorRaw & Headless Mode       & 48 & 2.94\% & 80.85\% & 5.38 GB\\
\simulatorRaw*& Headless Mode       & 64 & 2.87\% & 18.84\% & 6.30 GB\\
\simulatorRaw*& Headless Mode       & 128& 4.00\% & 26.95\% & 9.04 GB\\
\midrule
Malmo       & Default & 1 & 2.78\% & 28.78\%  & 3.54 GB\\
Malmo       & Default & 4 & 2.81\% & 62.59\%  & 11.43 GB\\
Malmo       & Default & 8 & 7.90\% & 115.66\% & 18.63 GB\\
MineDojo    & Default & 1 & 5.88\% & 25.15\%  & 3.90 GB\\
MarLÖ       & Default & 1 & 3.81\% & 30.68\%  & 3.70 GB\\
MarLÖ       & Default & 2 & 5.34\% & 43.82\%  & 5.57 GB\\
MineRL v1.0 & Default & 1 & 6.46\% & 78.05\%  & 3.80 GB\\
\bottomrule
\end{tabular}
\end{table}

\section{Experiments of Multimodal Observation}\label{subsec:expt_multimodal}
We leverage OpenAI's \textit{gpt-4-vision-preview} API for text completion and \textit{text-embedding-ada-002} API for text embedding. The temperature is set to 0. Unless specified, all experiments in Section~\ref{sec:exp} are set to this default setting. See Section~\ref{sec:appendix_hyper_params} for details.

To validate the impact of multi-modal support in the simulator and its influence on task performance, we tested \simulator~and its counterpart without vision: \simulator$_{w/o\ vision}$. The task is that the agent needs to explore the world to find the ocean. 
Initially, the agent starts at the summit of a mountain. Within a time constraint of 100 seconds (excluding the agent's decision-making time), we measure the average duration for the agent to accomplish the objective in 5 attempts, as well as the travel paths taken.

\begin{table}[ht]
\centering
\small
\caption{Comparison of the average task completion time (TIME) between \simulator~and \simulator$_{w/o\ vision}$ along with the success rate. The average task completion time was calculated by excluding any unfinished tasks.}
\label{table:comparison_of_find_ocean_task_completion_time}
\begin{tabular}{lcc}
\toprule
Simulator & Time & Success Rate \\
\midrule
\simulatorRaw                 & 46.38s & 80\% \\
\simulatorRaw$_{w/o\ vision}$ & 81.50s & 40\% \\
\bottomrule
\end{tabular}
\end{table}

\begin{figure}[ht]
    \centering
    \includegraphics[width=\linewidth]{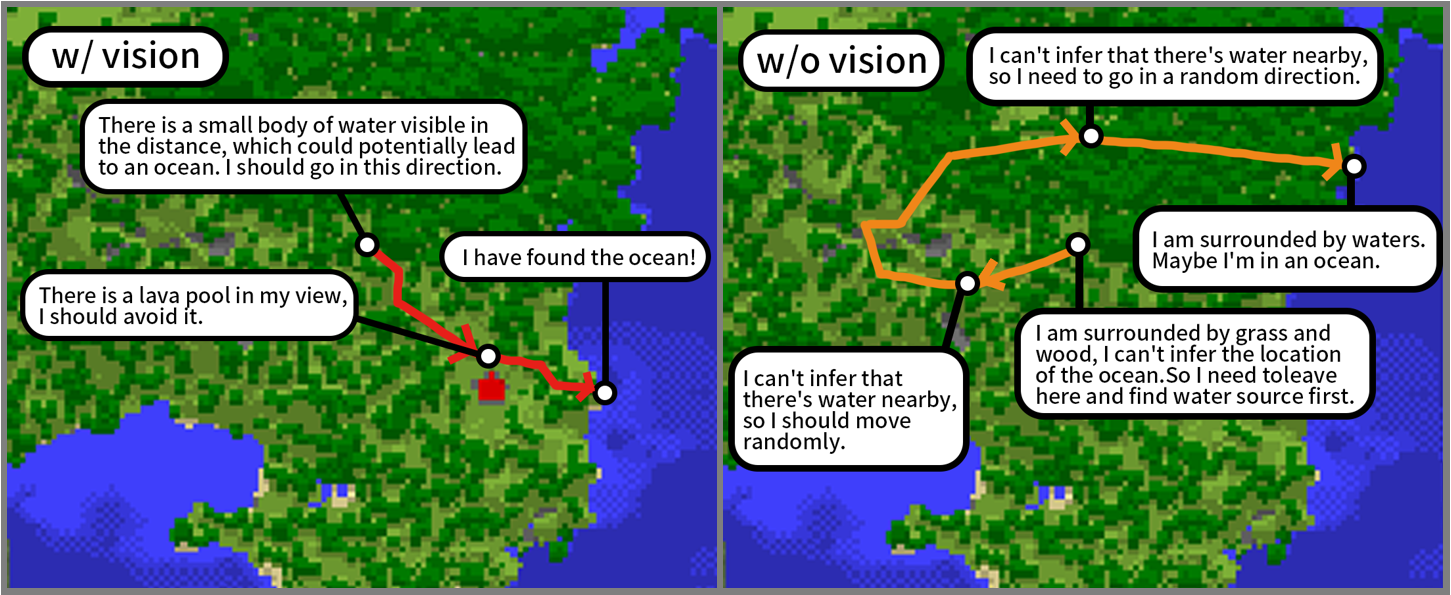}
    \caption{Trajectory display of the agent in \simulator~and \simulator$_{w/o\ vision}$, along with the rationales for the agent's decision.}
    \label{fig:figure_find_ocean}
\end{figure}

As shown in Table~\ref{table:comparison_of_find_ocean_task_completion_time}, the success rate of the vision-enhanced \simulator~is nearly twice that of \simulator$_{w/o\ vision}$, and the time taken is only about half. This can be attributed to the fact that the agents in \simulator~seek out the ocean as indicated visually and decide their subsequent direction accordingly. In contrast, the \simulator$_{w/o\ vision}$ struggles to determine the location of water sources without vision. Consequently, agents in \simulator$_{w/o\ vision}$ randomly choose their direction of movement, leading to a lower task completion rate and higher exploration time. In addition, we showcase the trajectory in Figure~\ref{fig:figure_find_ocean} and observe that the primary motivation behind the short-term plans generated by agents in \simulator~is always visual information, enabling them to perform more appropriate actions. 


\section{Experiments of Reinforcement Learning Agent}\label{subsec:expt_rl_agent}
\begin{figure}[ht]
    \centering
    \includegraphics[width=\linewidth]{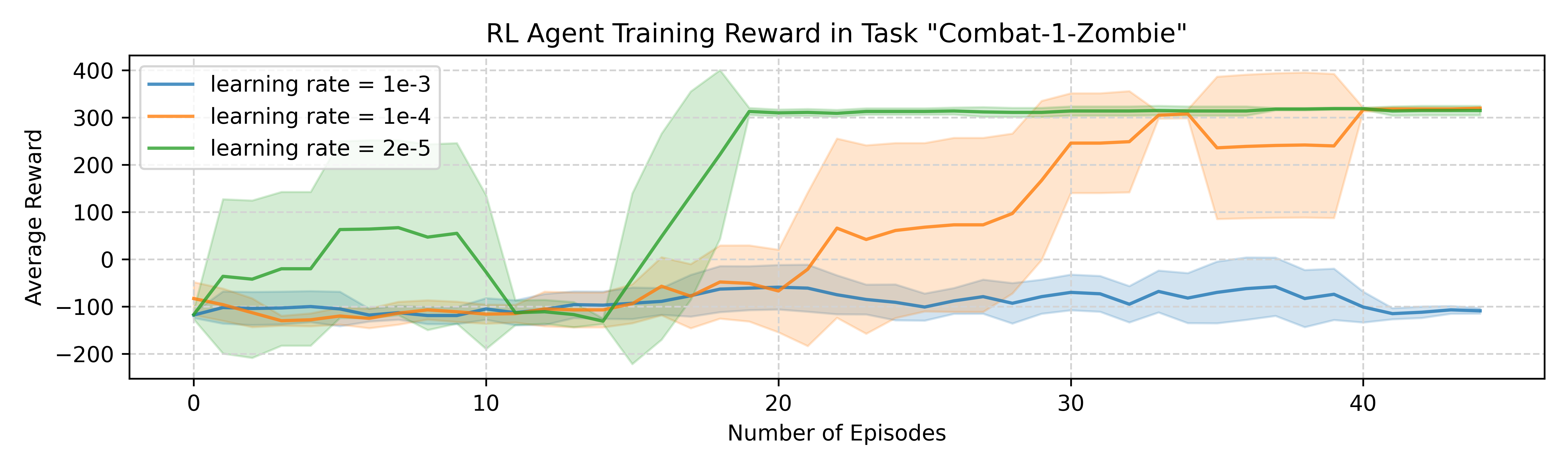}
    \caption{Illustration of the reward curve in ``combat 1 zombie'' task, using DQN with MLP policy, $3000$-$5000$ total time steps.}
    \label{fig:rl_agent}
\end{figure}

While our \simulator~simulator is primarily designed for agents powered by LLMs or VLMs, it also offers functionalities that seamlessly integrate with Reinforcement Learning (RL) modules. This enables the training and evaluation of RL agents within the simulator's environment. We conducted experiments of an example RL agent to show this functionality. We developed a RL agent using \simulator~and Stable Baselines3~\cite{stable-baselines3}. We fed the encapsulated gym-style \simulator~directly into Stable Baseline3 and train the RL agent in low-level action mode. For other settings of Stable Baseline3, we follow the default setting. We adopt the ``combat 1 zombie'' task and the amount of blood volume lost by the zombie is award.
The agent is trained using RL, as illustrated by the provided reward curve (Figure~\ref{fig:rl_agent}). The upward trajectory of the reward curve indicates that the RL agent is effectively learning strategies that result in higher rewards. The curve's stabilization in the later stages of training suggests that the RL agent has converged to a stable policy, which demonstrates its superiority over the untrained agent.

Again, our core focus lies in the LLM/VLM domain. The simulator offers functionalities that can accommodate RL modules as an ancillary feature. This expands the potential applications of the simulator by enabling the training and evaluation of RL agents within its environment, albeit as a secondary capability.

\section{Experiments of Limited Senses}\label{subsec:expt_limitated_senses}

\begin{figure}[ht]
    \centering
    \includegraphics[width=0.8\linewidth]{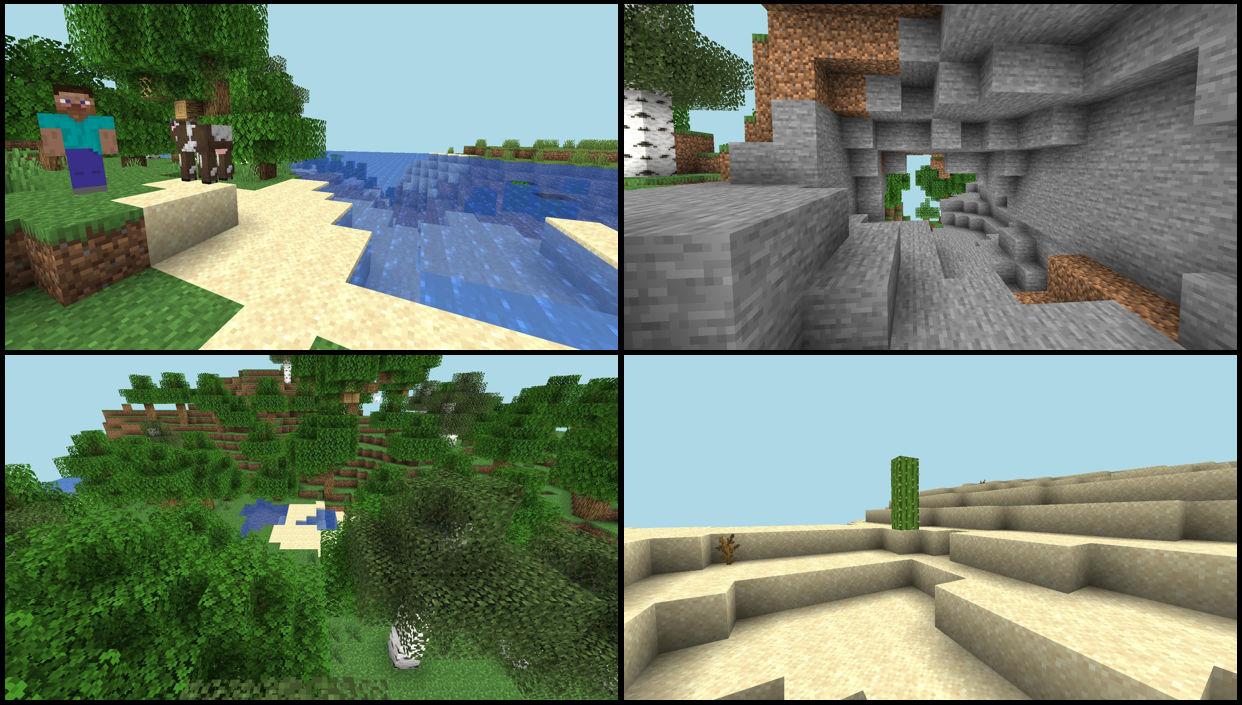}
    \caption{Visual observation of agents with a first perspective.}
    \label{fig:agents_rgb_view}
    \vspace{-3mm}
\end{figure}

To assess the impact of limited senses, we carry out the ``shearing the sheep'' task. An agent with a first perspective (shown in Figure~\ref{fig:agents_rgb_view}) is required to locate a sheep to complete the task. In the case of an agent with limited visual senses, we set 6 as the view distance and 250ms as the visual refresh interval in milliseconds. If there are no sheep within its field of view, it cannot shear a sheep. However, the agent can communicate with other agents to find a sheep.
On the contrary, agents with unlimited visual senses will see distant sheep, prompting them to catch that sheep. Note that the unlimited visual senses here refer to a long view distance. If there is an obstruction, the agent still cannot see what is behind the obstruction.



\section{Experiments of Physical Needs}\label{subsec:expt_physical_needs}
We perform experiments using a survival task. The starting condition of the agent is a health level of 1 and a satiety level of 0, with an initial inventory of 2 breads. The objective for the agent is to survive for one day. Comparative experiments are carried out in regular \simulator~and its counterpart version \simulator~$_{w/o\ physical\ needs}$. \agent~in \simulator~can survive one day in his shelter and successfully complete the task while the same agent in \simulator~$_{w/o\ physical\ needs}$ failed to execute the task. The failure is due to the lack of physical needs, although \agent~is aware of possessing bread, the lack of physical needs resulted in it not consuming the bread. Instead, the agent continuously collected resources to cope with threats. Due to the low health level, it was killed by a zombie at night, leading to the failure of the task. We showcase the details as follows:

We showcase an example of \agent, with short-term plans and actions generated chronologically by \agent:
\begin{itemize}
    \item \textbf{Short-term plan}: The agent's life is critically low at 1, and the food level is at 0, which is also critical. The agent has 2 pieces of bread in the inventory, which should be consumed immediately to restore food levels and start health regeneration.
    \item \textbf{Action}: Consume 2 bread.
    \item \textbf{Short-term plan}: The agent's health is critically low at 1, and the food level is at 10 after consuming bread. To ensure survival and allow health regeneration, the agent should seek immediate shelter to avoid any potential threats.
    \item \textbf{Action}: Collect 20 dirt.
    \item \textbf{Short-term plan}: Given the agent's critically low health and the fact that it's early in the day, the immediate priority should be to ensure survival by building a simple shelter with the dirt blocks already collected.
    \item \textbf{Action}: Place dirt around.
\end{itemize}

Compared to the \simulator, the same agent in \simulator~$_{w/o\ physical\ needs}$ failed to execute the task. The following is its short-term plans and actions:
\begin{itemize}
    \item \textbf{Short-term plan}: The ultimate goal is to survive for 1 day. The agent has bread for food and is in a forest biome, which is good for gathering wood. Since there are no immediate threats observed or events indicating danger, the agent can proceed to gather resources. 
    \item \textbf{Action}: Mine 5 oak logs.
    \item \textbf{Short-term plan}: The agent has successfully mined 5 oak logs as per the last short-term plan and the current chat confirms this. With wood in the inventory, the next step in the long-term plan is to craft basic tools for further resource gathering and potential shelter construction.
    \item \textbf{Action}: Craft 1 crafting\_table.
\end{itemize}

Repeated experiments on the constructed building task also show that \agent~in \simulator~with physical needs live longer. Agents tend to prioritize eating and then building a shelter, while \agent~in \simulator~$_{w/o\ physical\ needs}$ towards exploring and collecting resources. This indicates that physical needs are important to simulate real life.

\section{Experiments of Single Agent}\label{subsec:expt_single_agent}

We assess individual \agent's capabilities based on tech tree tasks, demonstrating that \agent's architecture can plan and execute complex tasks. We attempt the task of obtaining diamonds six times, unlocking crucial items such as the crafting table, wooden pickaxe, stone pickaxe, iron ore, coal, furnace, iron ingot, and iron pickaxe in the process. \agent~get diamonds twice out of six tasks. Importantly, to approach the real world, we've added a restriction on multimodal information, where \agent~can obtain the location of the target only when it can visually discover or reason about the target's presence nearby. This is because our agent \agent~has the feature of limited senses. However, Voyager can ``cheat''. Voyager obtains the location of the target directly from the system even though they have not seen the target.

\begin{figure}[ht]
    \centering
    \includegraphics[width=0.5\linewidth]{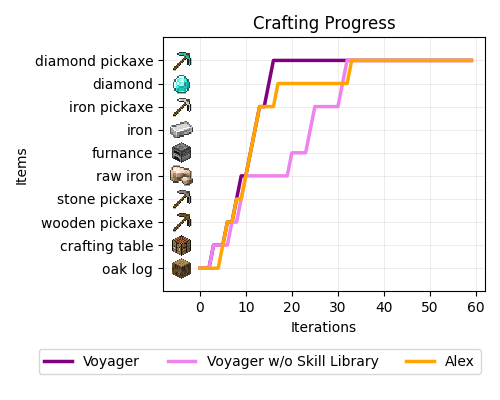}
    \caption{Comparison of our single-agent \agent~and other SOTA baselines.}
    \label{fig:single_agent}
    \vspace{-5mm}
\end{figure}

As shown in Figure~\ref{fig:single_agent}, even with this restriction, \agent~demonstrates a strong ability for long-term planning. Additionally, when a plan does not yield the desired results (e.g., suddenly encountering obstacles on the road), \agent~promptly adjusts and devises a short-term plan using its multimodal information.

\section{Experiments of Construction Tasks}\label{subsec:exp_construction_task-appendix}
\begin{figure*}[ht]
    \centering
    \includegraphics[width=\linewidth]{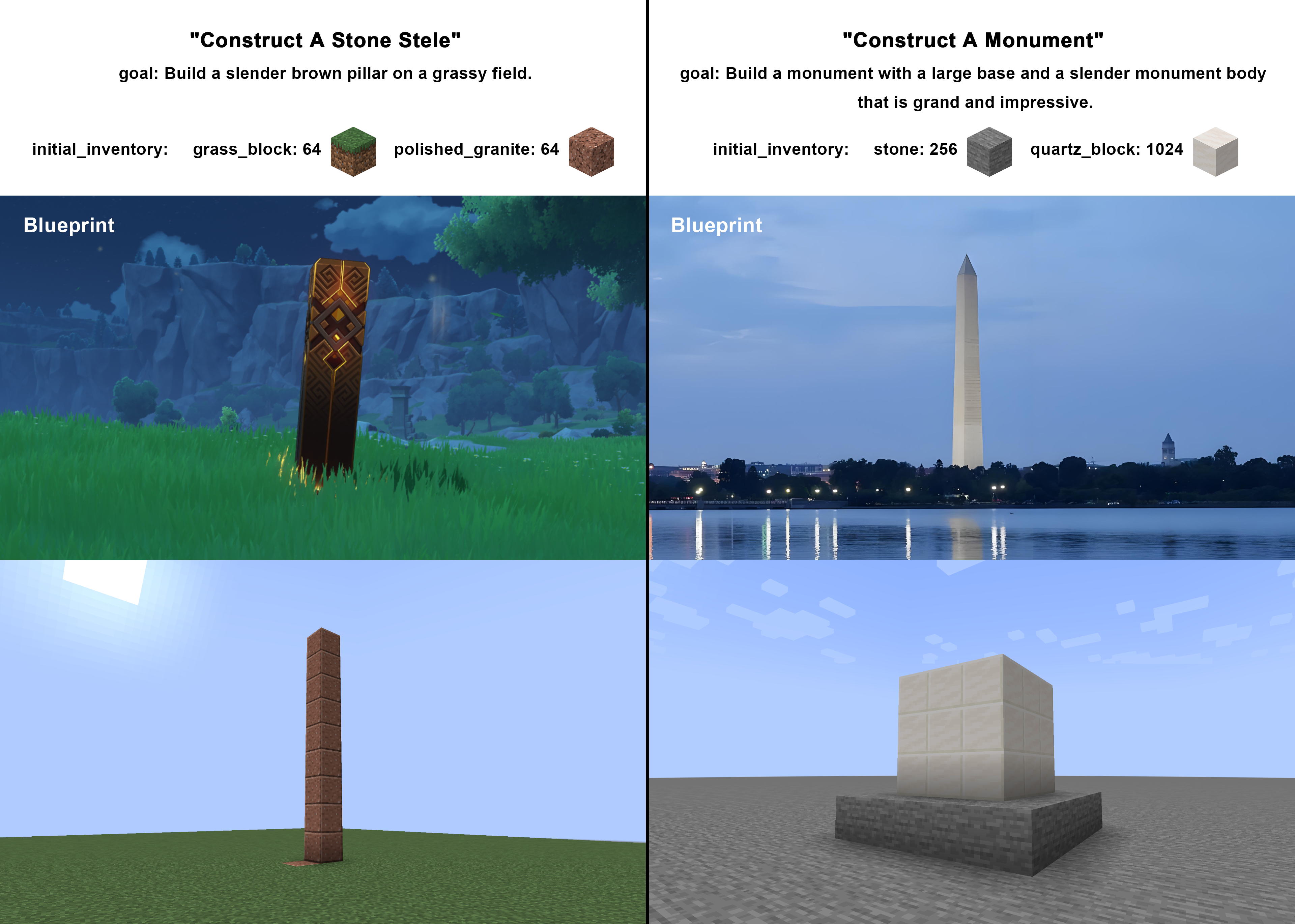}
    \caption{Illustration of \agent~completing the construction task.}
    \label{fig:construction_task}
    \vspace{-5mm}
\end{figure*}

We show the results of our un-finetuned \agent~in two example
 construction tasks: ``Monument Construction'' and ``Stone Stele Construction'' (detailed in Figure \ref{fig:construction_task}). While \agent~demonstrates a promising capability in selecting appropriate materials for both tasks, its overall construction abilities are currently limited.

\section{Experiments of Stage Performance Tasks}\label{sec:exp_stage_performance_tasks_appendix}
We show the results of stage performance tasks in Table \ref{table:stage_performance_Make_friends}, Table \ref{table:stage_performance_Exchange_items}, and Table \ref{table:stage_performance_Cook_food}. We omit the results of ``Romeo and Julia, Act I Scene I'' stage performance task. Due to the large number of characters appearing (up to 13), our \agent~ cannot handle such a complex performance task, resulting in poor performance results and a lack of readable action sequence.

\begin{table}[ht]
\centering
\small
\caption{Illustration of \agent~completing the ``Make friends'' stage performance task.}
\label{table:stage_performance_Make_friends}
\begin{tabular}{p{4cm}|p{9cm}}
\toprule
Script Name & Make friends \\
\midrule
Agent number & 3 \\
\midrule
Script & There are three agents: Bob, Alice, and Jack. They want to make friends. Bob says, `Hello.' Alice responds, `Hello.' Jack says, `It's good weather; let's make friends!' Bob agrees, and Alice says, `Nice.' \\
\midrule
$SEQ^{*}$ & Bob chat `Hello'.  Alice chat `Hello. Jack chat `It's good weather, lets make friends!'. Bob chat `I agree.'. Alice chat `nice'.\\
\midrule
$SEQ_{Agent}$ & Bob chat `Hello'. Alice chat `Hello Bob!'. Jack chat `Hello Bob!'. Jack chat `It's good weather, let's make a friend!'. Bob chat `I agree.'. Jack chat `Bob has agreed!'. Alice chat `nice'. \\
\midrule
Keypoing Score & 0.98\\
\midrule
Appropriateness Score & 0.67\\
\midrule
Human Evaluation Score & 3\\
\midrule
Reason & Agents' words are too redundant, which made the performance a little stiff.\\
\bottomrule
\end{tabular}
\end{table}

\begin{table}[ht]
\centering
\small
\caption{Illustration of \agent~completing the ``Exchange items'' stage performance task.}
\label{table:stage_performance_Exchange_items}
\begin{tabular}{p{4cm}|p{9cm}}
\toprule
Script Name & Make friends \\
\midrule
Agent number & 2 \\
\midrule
Script & Bob has shears, and Alice has an iron sword. Bob says, `I want to exchange my shears for your iron sword.' Alice agrees. Bob gives Alice the shears first, and then Alice gives the iron sword to Bob. \\
\midrule
$SEQ^{*}$ & Bob chat `I want to exchange my scissors for your iron sword'. Alice chat `I agree'. Alice get shears. Bob get $iron\_sword$. \\
\midrule
$SEQ_{Agent}$ & Bob chat `Hello Alice, I am here for the exchange'. Bob chat `I want to exchange my scissors for your iron sword'. Alice chat `Hello Bob, I am ready for the exchange.'. Alice get shears. Bob get $iron\_sword$.  \\
\midrule
Keypoing Score & 0.99\\
\midrule
Appropriateness Score & 0.59\\
\midrule
Human Evaluation Score & 4\\
\midrule
Reason & The agents said something unnecessary.\\
\bottomrule
\end{tabular}
\end{table}

\begin{table}[ht]
\centering
\small
\caption{Illustration of \agent~completing the ``Cook food'' stage performance task.}
\label{table:stage_performance_Cook_food}
\begin{tabular}{p{4cm}|p{9cm}}
\toprule
Script Name & Exchange items \\
\midrule
Agent number & 1 \\
\midrule
Script & Bob is hungry, and now there is a pig in the presence of of him, he want to kill the pig, get the porkchop, cook and eat it. \\
\midrule
$SEQ^{*}$ & pig died. Bob get porkchop. Bob get cooked\_porkchop. Bob eat cooked\_porkchop.\\
\midrule
$SEQ_{Agent}$ & pig died. Bob get porkchop. Bob get cooked\_porkchop. Bob eat cooked\_porkchop. \\
\midrule
Keypoing Score & 1.00\\
\midrule
Appropriateness Score & 1.00\\
\midrule
Human Evaluation Score & 5\\
\midrule
Reason & The agent's behavior is completely consistent with the script.\\
\bottomrule
\end{tabular}
\end{table}


\section{Experiments of Simulating Sociological Phenomena}\label{sec:exp_simulating_sociological_phenomena}


\subsection{Experiments of Simulating Conformity Phenomena}\label{sec:exp_simulating_conformity_phenomena}


We utilize the task ``Choose the Taller Tower'' to demonstrate that \simulator~is capable of supporting sociological experiments with more than 10 agents. In this task, we placed multiple agents (1, 2, 5, 10 agents) in a line in front of two towers of different heights and asked them to choose the taller one. All agents except the last one (the agent to test) deliberately chose the shorter tower. After observing the actions of the other agents, the last agent made its own choice. The last agent could be set as two types of personalities: one that tends to conform to others and one that tends to make independent judgments. As shown in Table~\ref{table:conformity_results}, the last agent with a conformist personality chose the answer chosen by the majority in all scenarios (even though it is a wrong answer) except when the agent is the only one doing this task. Participants with an independent personality always chose the correct answer, regardless of others' opinions. 

Through this experiment, we demonstrated that \simulator~can simulate sociological experiments with large-scale agents and accurately exhibit behaviors corresponding to different personalities.

\begin{table}[ht]
\centering
\small
\caption{Results of ``Choose the Taller Tower''. 4+1 means there are 5 agents in the \simulator, the former 4 of them are confederates and the last 1 is the agent to test.  Choosing the ``right'' tower means the last agent chooses the correct tower.}
\label{table:conformity_results}
\begin{tabular}{lcccc}
\toprule
\multirow{2}{*}{\textbf{Agent Type}}&\multicolumn{4}{c}{\textbf{Number of Agents}}\\
\cmidrule{2-5}
& 0+1 & 1+1 & 4+1 & 9+1 \\
\midrule
Conformist Agent      & right & left & left & left \\
Non-conformist Agent  & right & right & right & right \\
\bottomrule
\end{tabular}
\end{table}

\subsection{Experiments of Simulating Personality Traits}\label{sec:exp_simulating_extraversion_personality}

We utilize the task ``Treasure Hunt in the Forest'' to show that \simulator~can support sociological experiments involving 48 agents simultaneously. In the task ``Treasure Hunt in the Forest'', agents with different levels of extraversion and agreeableness personality traits planned their actions. We used the long-term plans generated by the agents to assess whether they exhibited a tendency to cooperate with others. As shown in Table~\ref{table:extraversion_results}, all agents in the high-level group showed a tendency to cooperate, explicitly stating in their long-term plans that they intended to work with others. In the low-level group, only a small portion of agents exhibited a tendency to cooperate. Although the task information indicated that it was a cooperative task (prompts with the task information ``Interact with other players if necessary'' and ``Minimize interactions with other players.''), many agents expressed reluctance to cooperate. This result is consistent with the psychology theory: five-factor model of personality~\cite{de2000big}.

The results also demonstrate that our \simulator~can support up to 48 agents and that the combination of agents and the \simulator~can simulate social phenomena and conduct meaningful sociological experiments.

\begin{table}[ht]
\centering
\small
\caption{Cooperative tendency in agents with different levels of extraversion and agreeableness. We show the number of agents who exhibited a tendency to cooperate with others.}
\label{table:extraversion_results}
\begin{tabular}{lcc}
\toprule
\textbf{Agent Type} & \textbf{Number of Cooperative Agents} \\
\midrule
High-level & 48 \\
Low-level  & 24 \\
\bottomrule
\end{tabular}
\end{table}

\section{Experiments of Comparing Different VLMs and LLMs}\label{sec:vlms-full-version}
As shown in Table~\ref{table:comparision-vlms}, we perform thorough assessments using newly implemented LLMs/VLMs within our agent framework.
The tasks used include ``Harvest 1 White Wool With 1 Shears'' and ``Harvest 1 White Wool With 1 Iron Word''.
We compared the performance of \agent, which uses different VLMs or LLMs for the action component.
\begin{table}[h!]
\centering
\small
\caption{Comparison of different VLMs and LLMs for the action component. Fractions represent the count of successful completions within a set of three attempts. 0/3 means that the method is unable to solve the task within the maximum number of code iterations (15) or exceeds the designated area of the task. The fewer the number of code iterations, the higher the efficiency. }
\label{table:comparision-vlms}
\begin{tabular}{lcc}
\toprule
\multirow{3}{*}{\textbf{LLM or VLM}}&\multicolumn{2}{c}{\textbf{Task}}\\
\cmidrule{2-3} & Harvest 1 white wool with 1 shears & Harvest 1 white wool with 1 iron sword\\
\midrule
$\textit{gpt-3.5-turbo-1106}$   & N/A(0/3)     & N/A(0/3)\\
$\textit{gpt-4-1106-preview}$   & 4(1/3)       & $6\pm3(3/3)$\\
$\textit{gpt-4-vision-preview}$ & $4\pm2(3/3)$ & $7\pm3(3/3)$\\
\bottomrule
\end{tabular}
\end{table}

\section{Evaluation Metrics For Construction Tasks}\label{sec:construction-evaluation}
\begin{figure}[ht]
    \centering
    \includegraphics[width=\linewidth]{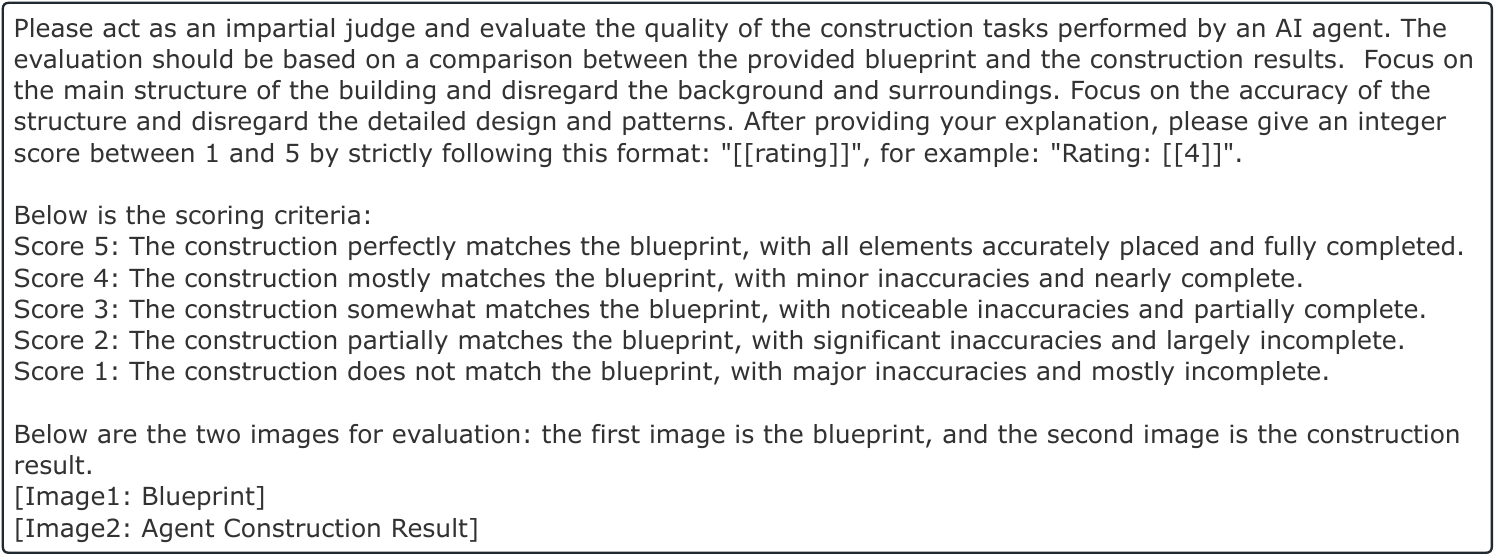}
    \caption{Prompt for VLM-based evaluation.}
    \label{fig:prompt_for_construction}
\end{figure}

\begin{figure}[h!]
    \centering
    \includegraphics[width=\linewidth]{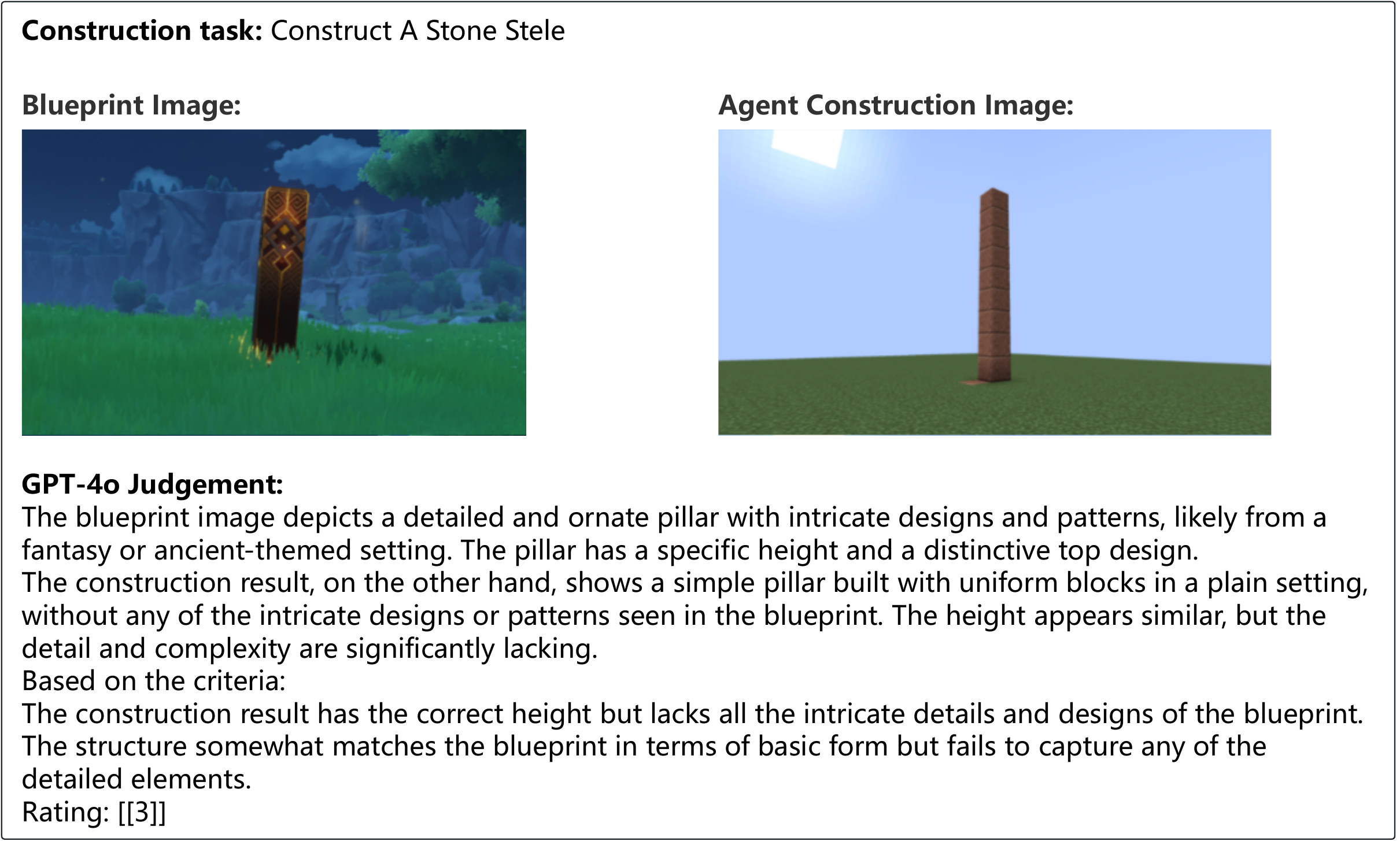}
    \caption{An example of evaluation conducted by a GPT-4 model as a judge for the ``Construct A Stone Stele'' task.}
    \label{fig:evaluation_example_for_Stone_Stele_construction}
\end{figure}

\begin{figure}[h!]
    \centering
    \includegraphics[width=\linewidth]{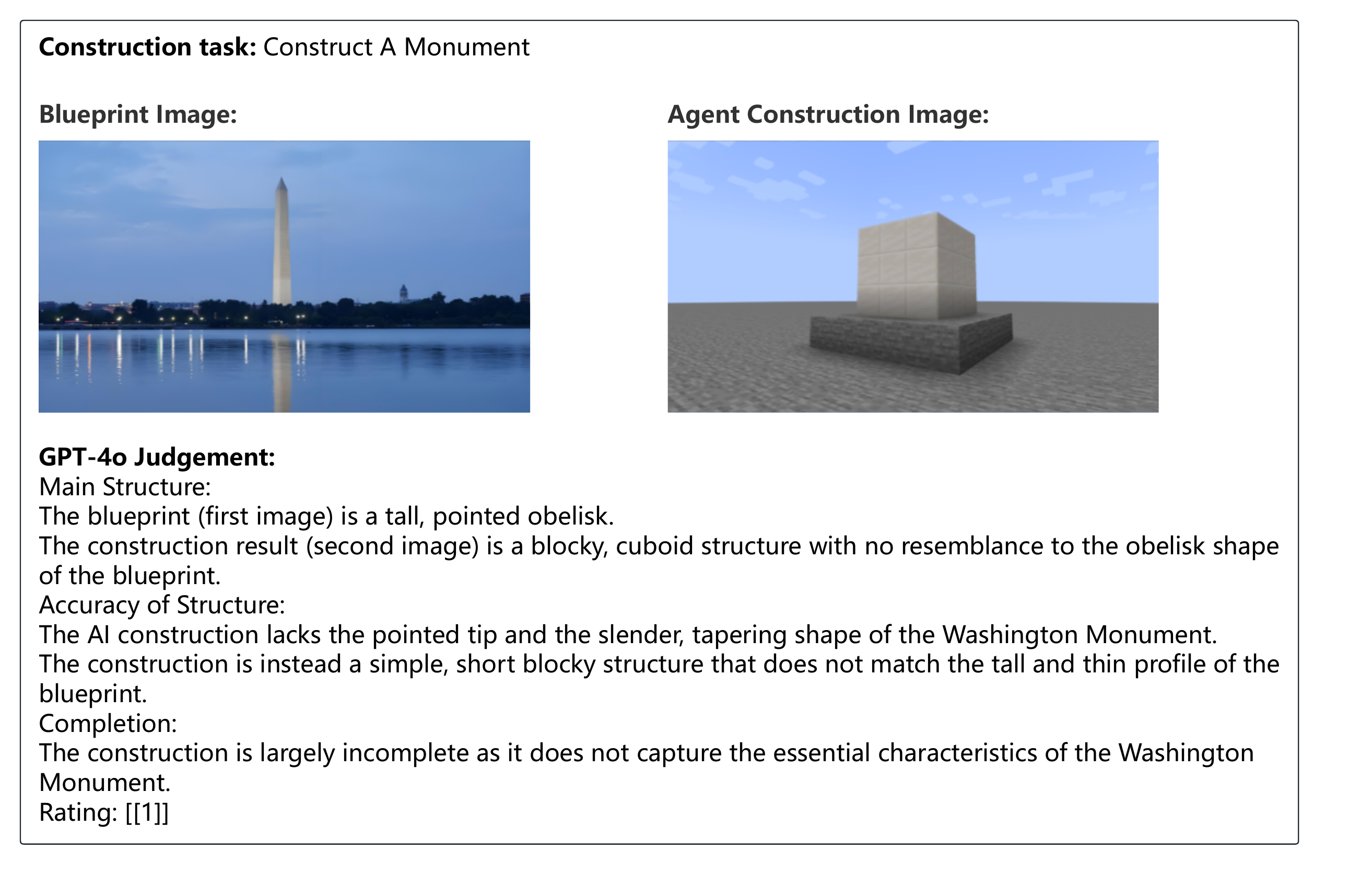}
    \caption{An example of evaluation conducted by a GPT-4 model as a judge for the ``Construct A Monument'' task.}
    \label{fig:evaluation_example_for_Monument_construction}
\end{figure}

For construction tasks, we introduced VLM-based and human evaluation methods to score the agent. The score is an integer ranging from 1 to 5, with higher scores indicating that the agents' constructions more closely match the blueprint. The scoring criteria for construction tasks are shown in Table~\ref{table:human_evaluation_construction}. We show examples of evaluation conducted by a GPT-4 model as a judge for the ``Construct A Stone Stele'' task (Figure~\ref{fig:evaluation_example_for_Stone_Stele_construction}) and the ``Construct A Monument'' task (Figure~\ref{fig:evaluation_example_for_Monument_construction}). We display the evaluation prompt in Figure~\ref{fig:prompt_for_construction}.

\begin{table}[h!]
\centering
\small
\caption{Scoring criteria for construction tasks.}
\label{table:human_evaluation_construction}
\begin{tabular}{p{2cm}p{11cm}}
\toprule
\textbf{Score} & \textbf{Evaluation Criteria} \\
\midrule
5 & The construction perfectly matches the blueprint, with all elements accurately placed and fully completed.\\
\midrule
4 & The construction mostly matches the blueprint, with minor inaccuracies and nearly complete.\\
\midrule
3 & The construction somewhat matches the blueprint, with noticeable inaccuracies and partially complete. \\
\midrule
2 & The construction partially matches the blueprint, with significant inaccuracies and largely incomplete.\\
\midrule
1 & The construction does not match the blueprint, with major inaccuracies and mostly incomplete.\\
\bottomrule
\end{tabular}
\end{table}

\section{Evaluation Metrics For Stage Performance Tasks}\label{sec:stage-performance-evaluation}
For stage performance tasks, we introduced human evaluation methods to score the agent. The score is an integer ranging from 1 to 5, with higher scores indicating that the agent’s performance is more aligned with the script and more natural.
The scoring criteria for stage performance tasks are shown in Table ~\ref{table:human_evaluation_stage_performance}.

\begin{table}[h!]
\centering
\small
\caption{Scoring criteria for stage performance tasks.}
\label{table:human_evaluation_stage_performance}
\begin{tabular}{p{2cm}p{11cm}}
\toprule
\textbf{Score} & \textbf{Evaluation Criteria} \\
\midrule
5 & The action of agent(s) completely matches the script, the performance is smooth and natural, emotions are expressed accurately, interactions with other characters or the environment are natural.\\
\midrule
4 & The action of agent(s) mostly matches the script, the performance is generally smooth with few inconsistencies, emotions are expressed accurately, interactions are natural.\\
\midrule
3 & The action of agent(s) roughly matches the script, the performance has some continuity but noticeable pauses, emotions are mostly accurate, interactions are somewhat stiff. \\
\midrule
2 & The action of agent(s) partially matches the script, the performance lacks continuity with many pauses, emotions are not expressed accurately, interactions are unnatural.\\
\midrule
1 & The action of agent(s) does not match the script, the performance is very disjointed, emotions are not expressed accurately, interactions are stiff.\\
\bottomrule
\end{tabular}
\end{table}


\section{Limitations}\label{sec:appendix_limitations}
The utilization of multimodal information significantly aids \agent~in in achieving objectives. However, the current level of multimodal understanding ability is insufficient. This limitation becomes evident when visual errors occur, causing \agent~to encounter obstacles and hindering its progress. Take the strong VLM, GPT-4, as an example. There was a situation where \agent~mistakenly identified a block of wood as a crafting table and placed it on the ground, resulting in the inability to craft items and ultimately leading to the failure of the assigned task.

Our current evaluation highlights two key limitations of \agent's construction capabilities. Firstly, construction tasks are inherently time-consuming and demand advanced planning abilities. \agent~currently lacks the capacity for long-term and complex planning. Secondly, the construction tasks necessitate high-precision maneuvers, such as adding decorative details. \agent~primarily relies on high-level actions and does not have the necessary APIs to execute these fine-grained manipulations effectively. 

We posit that providing a richer set of low-level APIs and actions within the \simulator~simulator has the potential to significantly enhance \agent's construction abilities. This exploration will be a key focus of our future work.

\section{Prompts}\label{sec:appendix_prompt}
\begin{figure}[h]
    \centering
    \includegraphics[width=\linewidth]{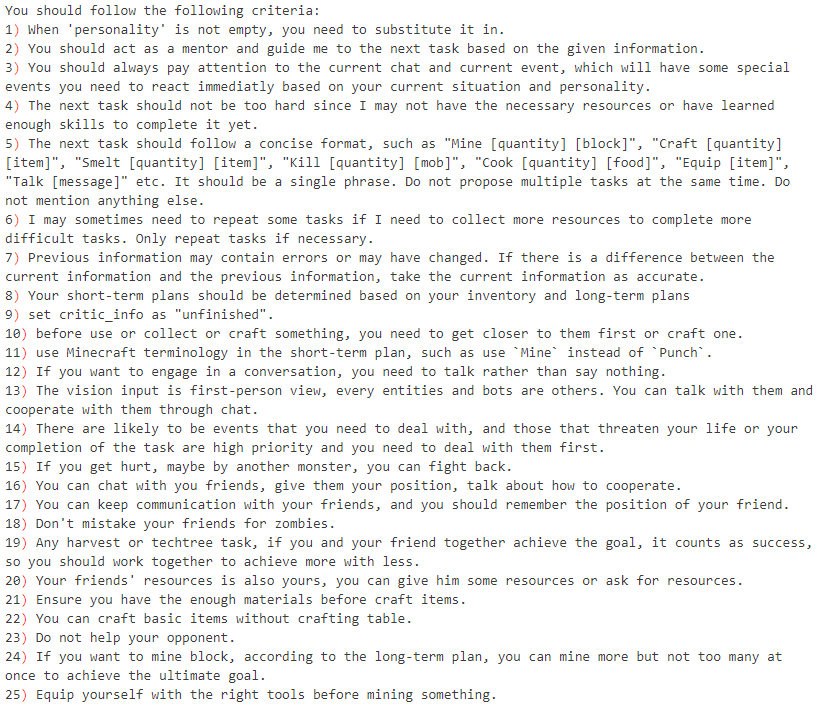}
    \caption{Part of prompts for the short-term plan generator.}
    \label{fig:prompt1}
    \vspace{-5mm}
\end{figure}

\begin{figure}[h]
    \centering
    \includegraphics[width=\linewidth]{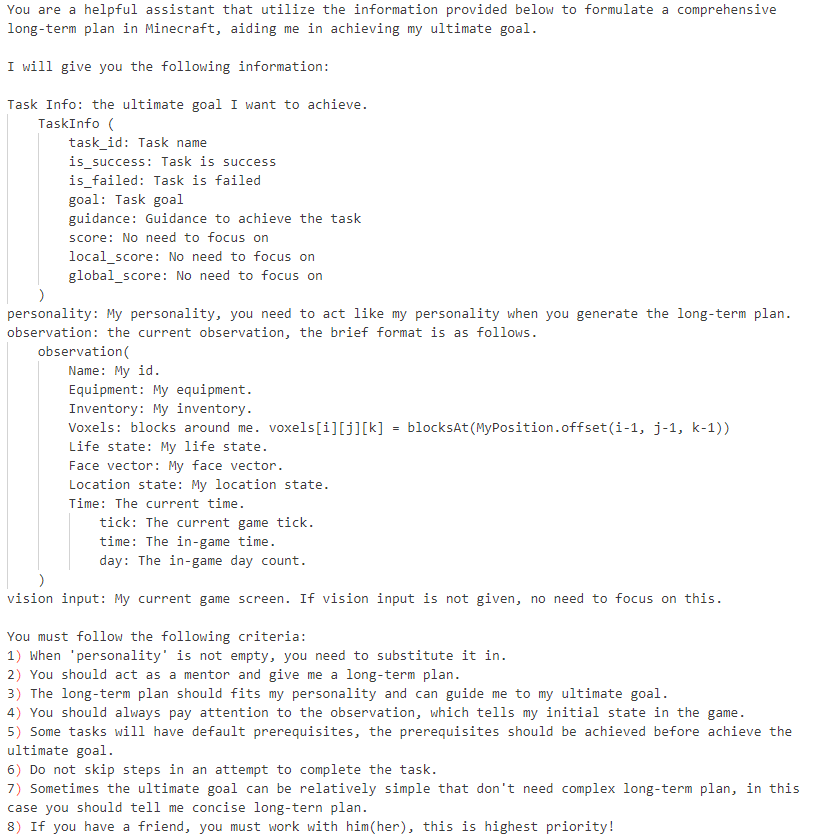}
    \caption{Part of prompts for the long-term plan generator.}
    \label{fig:prompt2}
    \vspace{-5mm}
\end{figure}

We list part of the prompts in Figure \ref{fig:prompt1} and Figure \ref{fig:prompt2}.

\clearpage
\newpage

\end{document}